\documentclass[nohyperref]{article}

\usepackage{microtype}
\usepackage{graphicx}
\usepackage{subcaption}
\usepackage{booktabs} 
\usepackage[table]{xcolor}
\usepackage{graphicx}
\usepackage{wrapfig}
\usepackage{tabularx}
\usepackage{enumitem}
\usepackage{tablefootnote}

\usepackage{hyperref}

\usepackage{booktabs}
\usepackage{multirow}

\usepackage{multicol}
\usepackage{xcolor}
\usepackage{nicefrac}

\usepackage[accepted]{icml2022}

\usepackage{amsmath}
\usepackage{amssymb}
\usepackage{mathtools}
\usepackage{amsthm}
\usepackage{tabularx}

\usepackage[capitalize,noabbrev]{cleveref}

\theoremstyle{plain}
\newtheorem*{theorem*}{Theorem}

\theoremstyle{definition}

\theoremstyle{remark}

\usepackage[textsize=tiny]{todonotes}
\usepackage{graphicx,caption}

\newcommand{\name} {Muse} 
\newcommand{\website}{muse-model.github.io}
	
\definecolor{LightCyan}{rgb}{0.88,1,1}
\definecolor{Grey}{rgb}{0.93,0.93,0.93}
\definecolor{DarkGrey}{rgb}{0.55,0.55,0.55}
\newcommand{\secc}[1]{Section \ref{sec:#1}}
\newcommand{\figg}[1]{Figure \ref{fig:#1}}
\newcommand{\tabb}[1]{Table \ref{tab:#1}}
\newcommand{\tablestyle}[2]{\setlength{\tabcolsep}{#1}\renewcommand{\arraystretch}{#2}\centering\footnotesize}
\newcolumntype{x}[1]{>{\centering\arraybackslash}p{#1pt}}
\newcolumntype{y}[1]{>{\raggedright\arraybackslash}p{#1pt}}
\newcolumntype{z}[1]{>{\raggedleft\arraybackslash}p{#1pt}}
\newlength\savewidth\newcommand\shline{\noalign{\global\savewidth\arrayrulewidth
  \global\arrayrulewidth 1pt}\hline\noalign{\global\arrayrulewidth\savewidth}}

\usepackage{fancyhdr}
\renewcommand{\headrulewidth}{0pt}
\fancypagestyle{empty}{\fancyfoot[C]{\vspace*{1.5\baselineskip}\thepage}}
\fancypagestyle{plain}{\fancyfoot[C]{\vspace*{1.5\baselineskip}\thepage}}
\newcommand{\mask}{{\texttt{[MASK]}}}

\newcommand{\lowres}{256}
\newcommand{\lowressq}{$\lowres \times \lowres$}
\newcommand{\highres}{512}
\newcommand{\highressq}{$\highres \times \highres$}

\newcommand{\ccfidsr}{6.06} 
\newcommand{\cocofid}{7.88} 
\newcommand{\cococlip}{0.32}

\begin{document}

\onecolumn
\icmltitle{\name: Text-To-Image Generation via Masked Generative Transformers}

\pagestyle{plain}



\icmlsetsymbol{equal}{*}
\icmlsetsymbol{core}{\textdagger}

\begin{icmlauthorlist}
\icmlauthor{Huiwen Chang}{equal}
\icmlauthor{Han Zhang}{equal}
\icmlauthor{Jarred Barber}{core}
\icmlauthor{AJ Maschinot}{core}
\icmlauthor{Jos\'e Lezama}{}
\icmlauthor{Lu Jiang}{}
\icmlauthor{Ming-Hsuan Yang}{}
\icmlauthor{Kevin Murphy}{}
\icmlauthor{William T. Freeman}{}
\icmlauthor{Michael Rubinstein}{core}
\icmlauthor{Yuanzhen Li}{core}
\icmlauthor{Dilip Krishnan}{core}
\end{icmlauthorlist}
\vspace{-5pt}
\begin{center}
    \large{Google Research}
\end{center}

\icmlcorrespondingauthor{Huiwen Chang}{huiwenchang@google.com}
\icmlcorrespondingauthor{Han Zhang}{zhanghan@google.com}
\icmlcorrespondingauthor{Dilip Krishnan}{dilipkay@google.com}

\icmlkeywords{Machine Learning, ICML}

\vskip 0.3in



\printAffiliationsAndNotice{\icmlEqualContribution $^\dagger$Core contribution} 

\begin{abstract}
We present \name, 
a text-to-image Transformer model that achieves state-of-the-art image generation performance while being significantly more efficient than diffusion or autoregressive models.
%
\name~is trained on a masked modeling task in discrete token space: given the text embedding extracted from a pre-trained large language model (LLM), \name~is trained to predict randomly masked image tokens.
%
Compared to pixel-space diffusion models, such as Imagen and DALL-E~2, \name~is significantly more efficient due to the use of discrete tokens and requiring fewer sampling iterations; 
compared to autoregressive models, such as Parti, \name~is more efficient due to the use of parallel decoding.
The use of a pre-trained LLM enables fine-grained language understanding, translating to high-fidelity image generation and the understanding of visual concepts such as objects, their spatial relationships, pose, cardinality etc.
%
Our 900M parameter model achieves a new SOTA on CC3M, with an FID score of \ccfidsr. 
The \name~3B parameter model achieves an FID of \cocofid~on zero-shot COCO evaluation, along with a CLIP score of \cococlip. 
Muse also directly enables a number of image editing applications without the need to fine-tune or invert the model: inpainting, outpainting, and mask-free editing. More results are available at \url{http://\website}.
%
\end{abstract}


\section{Introduction}


Generative image models conditioned on text prompts have taken an enormous leap in quality and flexibility in the last few years \citep{dalle2,glide,imagen,parti,ldm,midjourney}. This was enabled by a combination of deep learning architecture innovations \citep{vqvae,vaswani2017attention}; novel training paradigms such as masked modeling for both language \citep{bert,t5xxl} and vision tasks \citep{MAE,maskgit}; new families of generative models such as diffusion \citep{ddpm,ldm,imagen} and masking-based generation \citep{maskgit}; and finally, the availability of large scale image-text paired datasets \citep{laion}. 

\newcommand{\figwidth}{1.0\textwidth}

\begin{figure*}[ht!]
\vspace{-15pt}
\centering
\captionsetup{width=\figwidth}
\includegraphics[width=\figwidth]{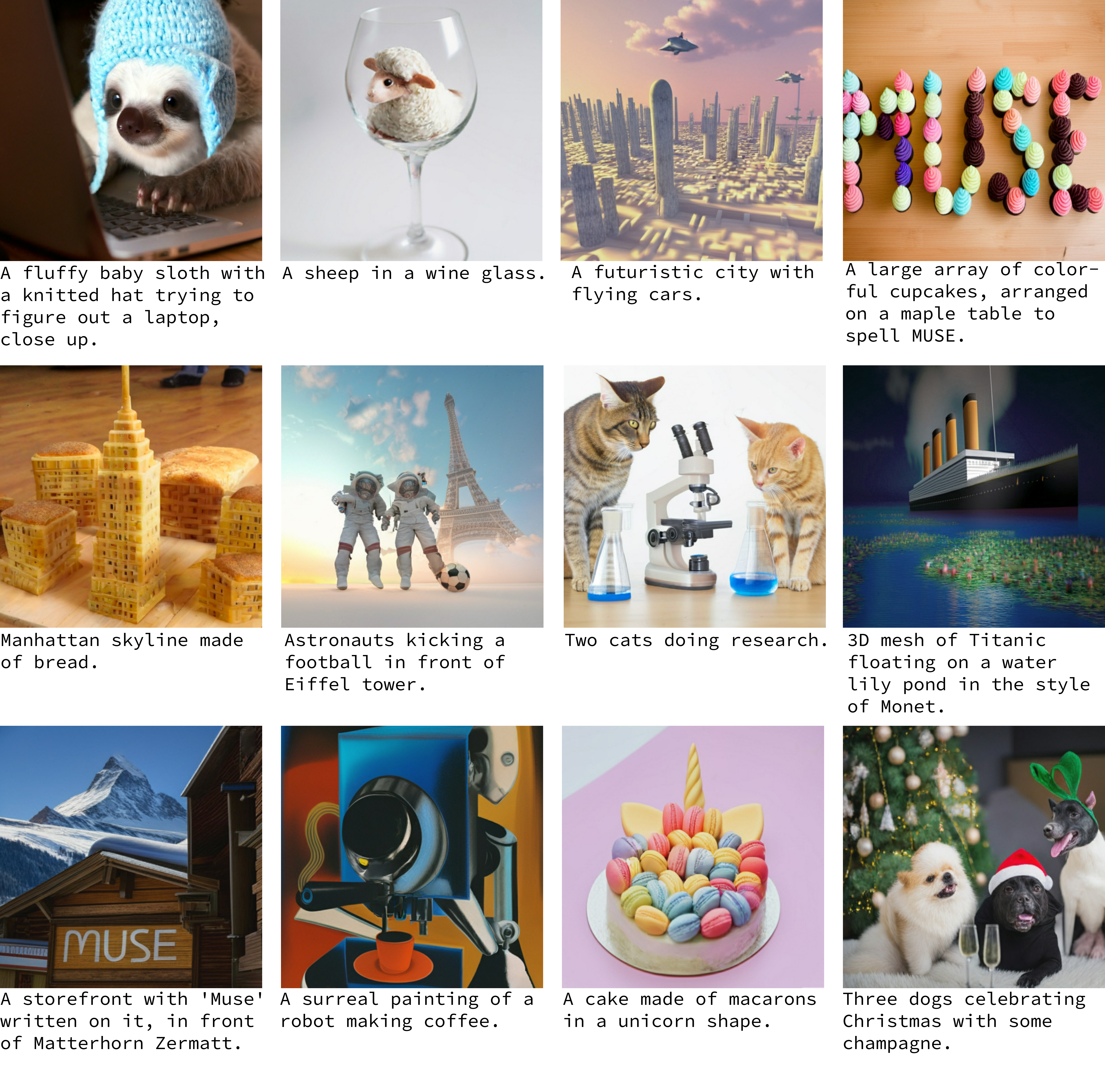}
\vspace{-23pt}
\caption{\small \name~text-to-image generation ($512 \times 512$ resolution). Under each generated image, the corresponding caption is shown, exhibiting a variety of styles, captions and understanding. Each image was generated in $1.3$s on a TPUv4 chip. 
}
\label{fig:teaser_t2i}
\end{figure*}

\renewcommand{\figwidth}{1.0\textwidth}
\begin{figure*}[ht!]
\centering
\captionsetup{width=\figwidth}
\includegraphics[width=\figwidth]{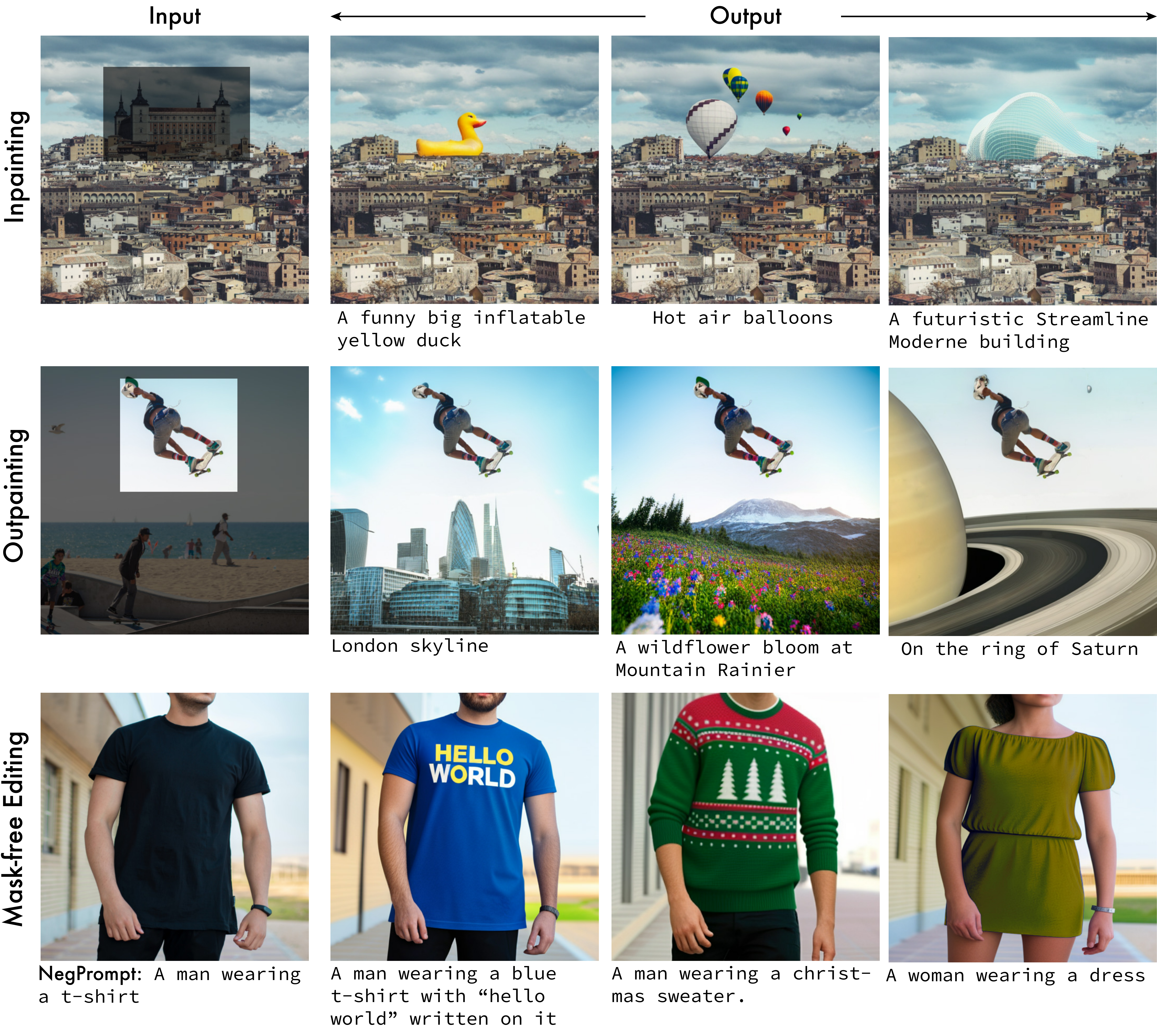}
\vspace{-18pt}
\caption{\small Examples of zero-shot text-guided image editing using \name. We show examples of a number of editing applications using the \name ~text-to-image generative model, on \emph{real} input images, without fine-tuning. All edited images are generated at $512\times512$ resolution. 
}
\vspace{-5pt}
\label{fig:teaser_edit}
\end{figure*}

In this work, we present a new model for text-to-image synthesis using a masked image modeling approach \citep{maskgit}. Our image decoder architecture is conditioned on embeddings from a pre-trained and frozen T5-XXL \citep{t5xxl} large language model (LLM) encoder. In agreement with Imagen \citep{imagen}, we find that conditioning on a pre-trained LLM is crucial for photorealistic, high quality image generation. Our models (except for the VQGAN quantizer) are built on the Transformer \citep{vaswani2017attention} architecture. 

\begin{figure*}[ht!]
\begin{center}
\vspace{-10pt}
\includegraphics[width=.85\textwidth]{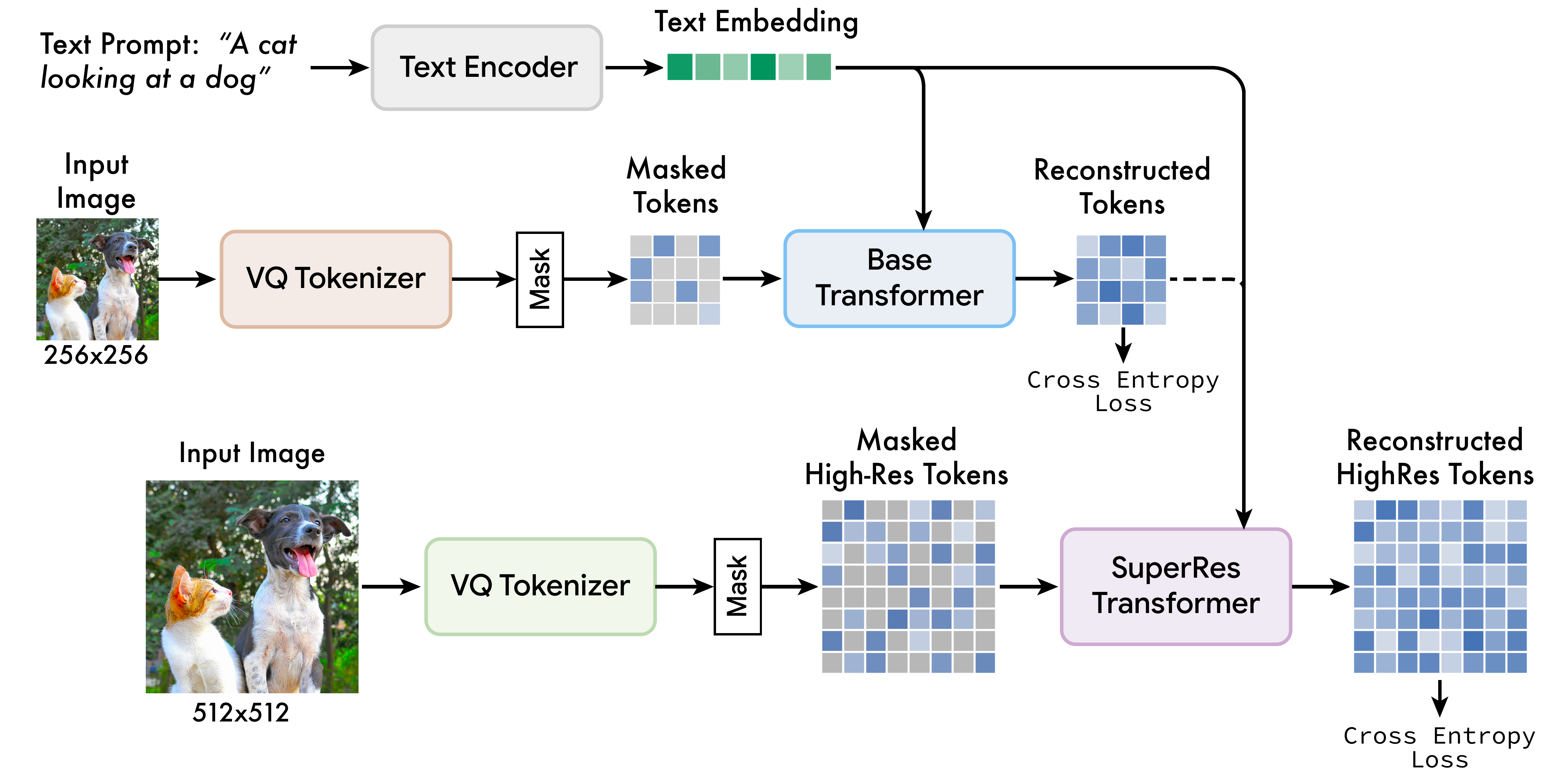}
\end{center}
\vspace{-15pt}
\caption{\small \name~Framework: We show the training pipeline for our model, with the T5-XXL pre-trained text encoder, base model and super-resolution model depicted on the three rows. The text encoder generates a text embedding that is used for cross-attention with image tokens for both base and super-res Transformer layers. The base model uses a VQ Tokenizer that is pre-trained on lower resolution ($\lowres\times\lowres$) images and generates a $16\times16$ latent space of tokens. This sequence is masked at a variable rate per sample and then the cross-entropy loss learns to predict the masked image tokens. Once the base model is trained, the reconstructed lower-resolution tokens and text tokens are passed into the super-res model that then learns to predict masked tokens at a higher resolution. }

\vspace{-10pt}
\label{fig:model}
\end{figure*}

We have trained a sequence of \name~models, ranging in size from 632M parameters to 3B parameters (for the image decoder; the T5-XXL model has an additional 4.6B parameters). Each model consists of several sub-models (\figg{model}):
First, we have a pair of VQGAN ``tokenizer'' models \citep{esser2021taming}, which can encode an input image to a sequence of discrete tokens as well as decode a token sequence back to an image. We use two VQGANs, one for 256x256 resolution (``low-res'') and another for 512x512 resolution (``high-res''). Second, we have a base masked image model, which contains the bulk of our parameters. This model takes a sequence of partially masked low-res tokens and predicts the marginal distribution for each masked token, conditioned on the unmasked tokens and a T5XXL text embedding. Third, we have a ``superres'' transformer model which translates (unmasked) low-res tokens into high-res tokens, again conditioned on T5-XXL text embeddings. We explain our pipeline in detail in \secc{model}.


Compared to Imagen \citep{imagen} or Dall-E2 \citep{dalle2} which are built on cascaded pixel-space diffusion models, \name~is significantly more efficient due to the use of discrete tokens; it can be thought of as a discrete diffusion process with the absorbing state (\mask)~\citep{austin2021structured}. Compared to Parti \citep{parti}, a state-of-the-art autoregressive model, \name~is more efficient due to the use of parallel decoding.
Based on comparisons on similar hardware (TPU-v4 chips), we estimate that \name~is more than $10$x faster at inference time than either Imagen-3B or Parti-3B models and $3$x faster than Stable Diffusion v1.4 \citep{ldm} (see \secc{speed}). All these comparisons are when images of the same size: either $256\times256$ or $512\times512$. \name~is also faster than Stable Diffusion \citep{ldm}, in spite of both models working in the latent space of a VQGAN. We believe that this is due to the use of a diffusion model in Stable Diffusion v1.4 which requires a significantly higher number of iterations at inference time.

The efficiency improvement of \name, however, does \emph{not} come at a loss of generated image quality or semantic understanding of the input text prompt. We evaluate our output on multiple criteria, including CLIP score \citep{clip} and FID \citep{fid}. The former is a measure of image-text correspondence; and the latter a measure of image quality and diversity. Our 3B parameter model achieves a CLIP score of \cococlip~and an FID score of \cocofid~on the COCO \citep{coco} zero-shot validation benchmark, which compares favorably with that of other large-scale text-to-image models (see \tabb{eval_coco}). 
Our 632M(base)+268M(super-res) parameter model achieves a state of the art FID score of \ccfidsr~ when trained and evaluated on the CC3M \citep{sharma2018conceptual} dataset, which is significantly lower than all other reported results in the literature (see \tabb{eval_cc3m}). We also evaluate our generations on the PartiPrompts \citep{parti} evaluation suite with human raters, who find that \name~generates images better aligned with its text prompt $2.7$x more often than Stable Diffusion v1.4 \citep{ldm}. 

\name~generates images that reflect different parts of speech in input captions, including nouns, verbs and adjectives.
Furthermore, we present evidence of multi-object properties understanding, such as compositionality and cardinality, as well image style understanding.
See \figg{teaser_t2i} for a number of these examples and our website \url{http://\website} for more examples. The mask-based training of \name~lends itself to a number of zero-shot image editing capabilities. A number of these are shown in \figg{teaser_edit}, including zero-shot, text-guided inpainting, outpainting and mask-free editing. More details are in \secc{results}. Our contributions are: 
\begin{enumerate}[topsep=1pt, partopsep=1pt, itemsep=1pt,parsep=1pt]
    \item We present a state-of-the-art model for text-to-image generation which achieves excellent FID and CLIP scores (quantitative measures of image generation quality, diversity and alignment with text prompts).
    \item Our model is significantly faster than comparable models due to the use of quantized image tokens and parallel decoding.
    \item Our architecture enables out-of-the-box, zero-shot editing capabilities including inpainting, outpainting, and mask-free editing.
\end{enumerate}


\section{Model}
\label{sec:model}
Our model is built on a number of components. Here, we provide an overview of each of those components in the order of their training, while relegating many details of the architecture and parameters to the Appendix. \figg{model} provides an overview of the model architecture. 

\subsection{Pre-trained Text Encoders}
Similar to the findings in \citep{imagen}, we find that leveraging a pre-trained large language model (LLM) is beneficial to high-quality image generation. The embeddings extracted from an LLM such as T5-XXL \citep{t5xxl} carry rich information about objects (nouns), actions (verbs), visual properties (adjectives), spatial relationships (prepositions), and other properties such as cardinality and composition. Our hypothesis is that the \name~model learns to map these rich visual and semantic concepts in the LLM embeddings to the generated images; it has been shown in recent work \citep{merullo2022linearly} that the conceptual representations learned by LLM's are roughly linearly mappable to those learned by models trained on vision tasks. Given an input text caption, we pass it through the frozen T5-XXL encoder, resulting in a sequence of 4096 dimensional language embedding vectors. These embedding vectors are linearly projected to the hidden size of our Transformer models (base and super-res).

\subsection{Semantic Tokenization using VQGAN}
\label{sec:vqgan}
A core component of our model is the use of semantic tokens obtained from a VQGAN \cite{esser2021taming} model. This model consists of an encoder and an decoder, with a quantization layer that maps an input image into a sequence of tokens from a learned codebook. We build our encoder and decoder entirely with convolutional layers to support encoding images from different resolutions. The encoder has several downsampling blocks to reduce the spatial dimension of the input, while the decoder has the corresponding number of upsampling blocks to map the latents back into original image size. Given an image of size $H \times W$, the encoded token is of size $\nicefrac{H}{f} \times \nicefrac{W}{f}$, with downsampling ratio $f$.
We train two VQGAN models: one with downsampling ratio $f=16$ and the other with downsampling ratio $f=8$. We obtain tokens for our base model using the $f=16$ VQGAN model on {\lowres}$\times${\lowres} pixel images, thus resulting in tokens with spatial size $16 \times 16$. We obtain the tokens for our super-resolution model using the $f=8$ VQGAN model on $512\times512$ images, and the corresponding token has spatial size $64\times64$. As mentioned in previous work \citep{esser2021taming}, the resulting discrete tokens after encoding capture higher-level semantics of the image, while ignoring low level noise. Furthermore, the discrete nature of these tokens allows us to use a cross-entropy loss at the output to predict masked tokens in the next stage.


\subsection{Base Model}

Our base model is a masked transformer\citep{vaswani2017attention,bert}, where the inputs are the projected T5 embeddings and image tokens. We leave all the text embeddings unmasked and randomly mask a varying fraction of image tokens  (see \secc{masking}) and replace them with a special \mask token \citep{maskgit}.
We then linearly map image tokens into image input embeddings of the required Transformer input/hidden size along with learned 2D positional embeddings. Following previous transformer architecture \citep{vaswani2017attention},  we use several transformer layers including self-attention block, cross-attention block and MLP block to extract features. At the output layer, an MLP is used to convert each masked image embedding to a set of logits (corresponding to the VQGAN codebook size) and a cross-entropy loss is applied with the ground truth token label as the target. At training, the base model is trained to predict all masked tokens at each step. However, for inference, mask prediction is performed in an iterative manner which significantly increases quality. See \secc{iterativedec} for details.

\subsection{Super-Resolution Model}

\begin{figure*}[ht!]
\vspace{-10pt}
\begin{minipage}[c]{0.63\textwidth}
\centering
\includegraphics[width=\textwidth]{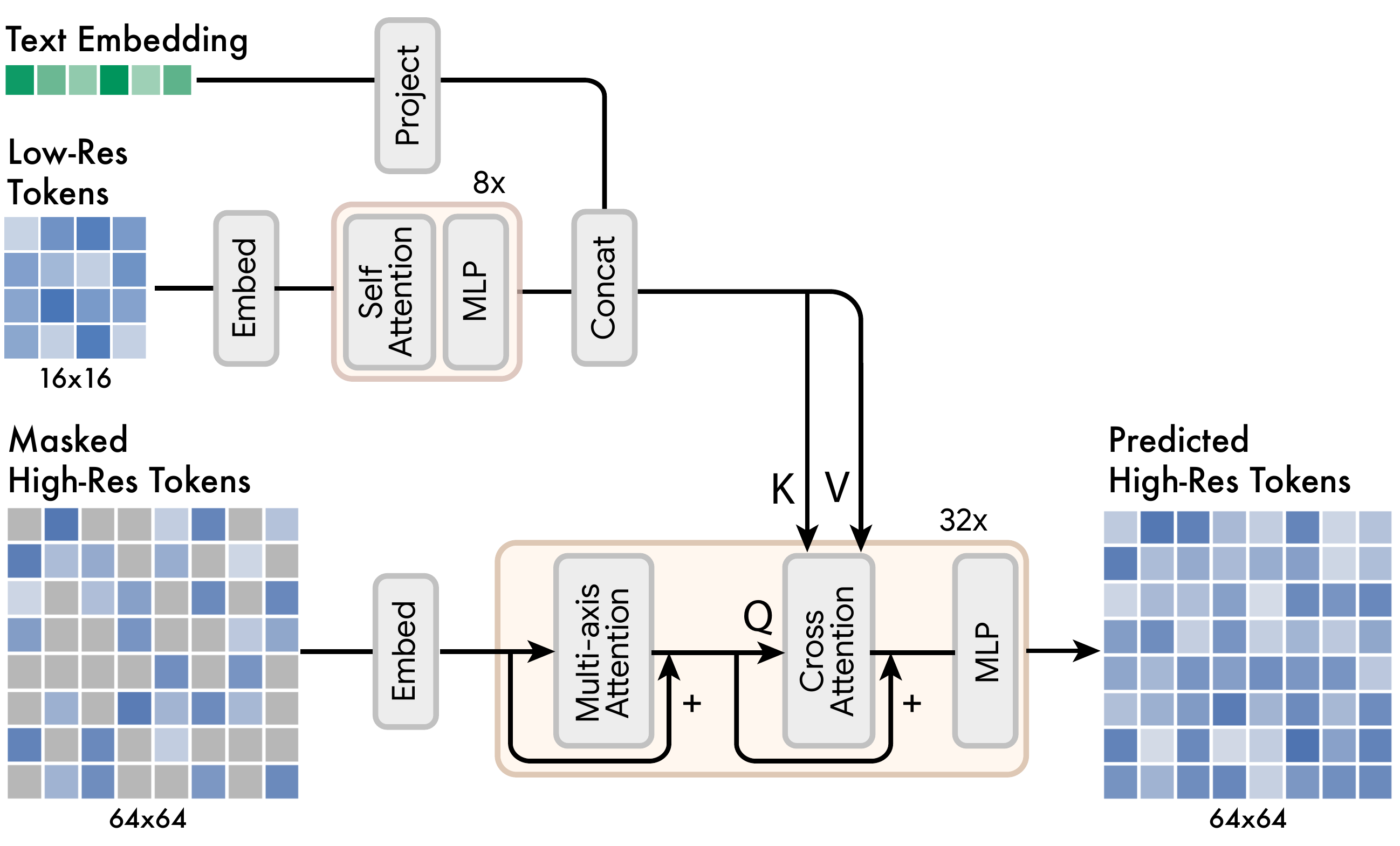}
\end{minipage} \hfill
\begin{minipage}[c]{0.34\textwidth}
\includegraphics[width=\textwidth]{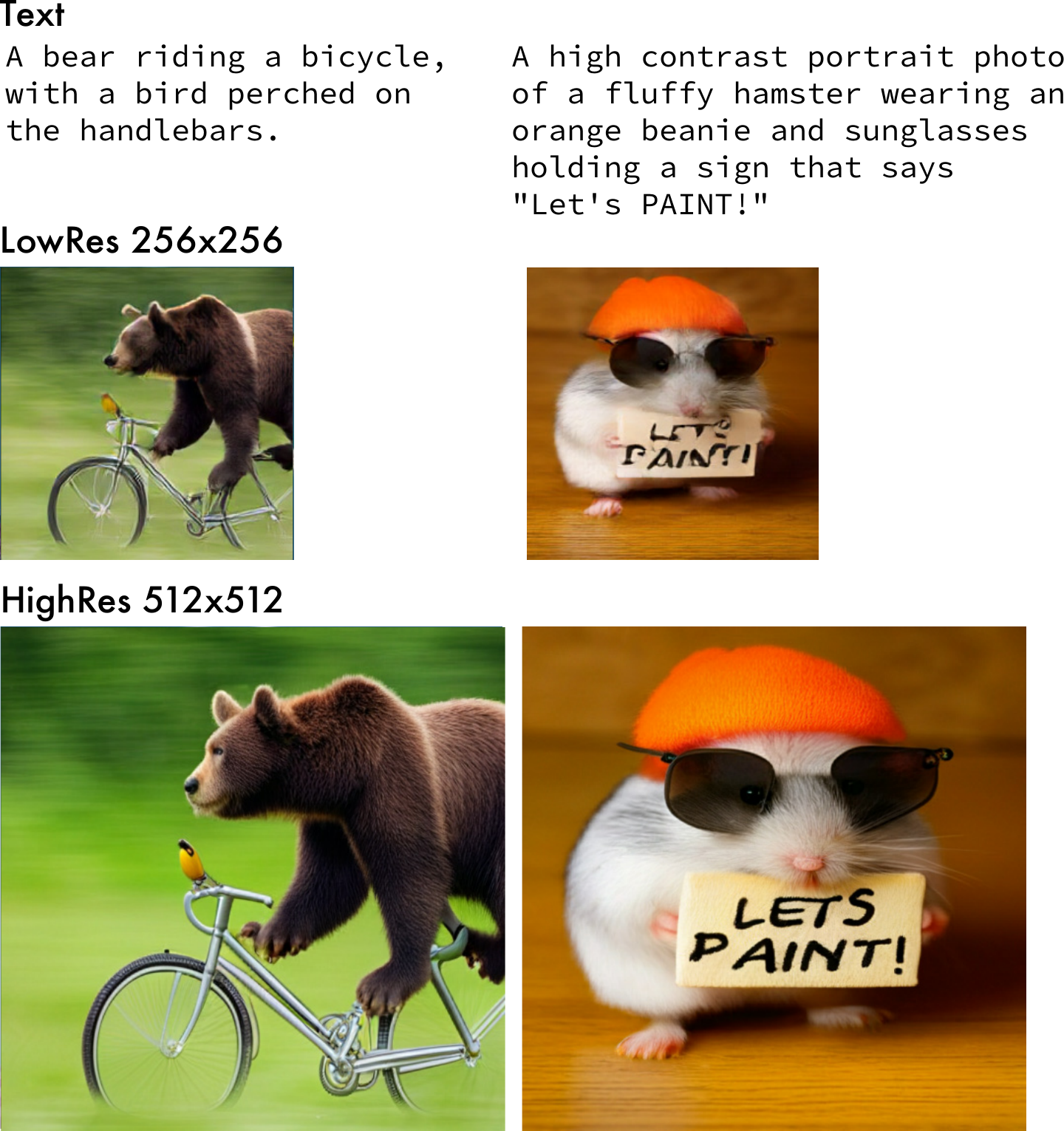}
\end{minipage} 
\vspace{-5pt}
\caption{\small Super-resolution Model. On the left is shown the architecture of the super-resolution model. Low-resolution tokens are passed into a series of self-attention Transformer layers; and the resulting output embeddings are concatenated with text embeddings extracted from the conditioning text prompt. Following this, cross-attention is applied from these concatenated embeddings to the masked high-resolution tokens; the loss learns to predict these masked tokens conditioned on the low-resolution and text tokens. On the right are shown two examples of the improvement brought about by the super-resolution model.}

\label{fig:sr}
\end{figure*} 
We found that directly predicting $512\times512$ resolution leads the model to focus on low-level details over large-scale semantics. As a result we found it beneficial to use a cascade of models: first a base model that generates a $16\times16$ latent map (corresponding to a $\lowres \times \lowres$ image), followed by a super-resolution model that upsamples the base latent map to a $64\times64$ latent map (corresponding to a $512\times512$ image). The super-res model is trained after the base model has been trained.

As mentioned in \secc{vqgan}, we trained two VQGAN models, one at $16\times16$ latent resolution and $\lowres\times\lowres$ spatial resolution, and the second at $64\times64$ latent resolution and $512\times512$ spatial resolution. Since our base model outputs tokens corresponding to a $16\times16$ latent map, our super-resolution procedure learns to ``translate" the lower-resolution latent map to the higher-resolution latent map, followed by decoding through the higher-resolution VQGAN to give the final high-resolution image. This latent map translation model is also trained with text conditioning and cross-attention in an analogous manner to the base model, as shown in \figg{sr}. 

\subsection{Decoder Finetuning}
\label{sec:dec_finetune}
To further improve our model's ability to generate fine details, we increase the capacity of the VQGAN decoder by the addition of more residual layers and channels while keeping the encoder capacity fixed. We then finetune the new decoder layers while keeping the VQGAN encoder weights, codebook and transformers (i.e., base model and super resolution model) frozen. This allows us to improve our visual quality without re-training any of the other model components (because the visual token ``language'' stays fixed). This is shown in \figg{finetune_decoder} in the Appendix, where we see that the finetuned decoder can reconstruct more sharper details in the store front. We also give details of the finetuned decoder architecture in the Appendix.

\subsection{Variable Masking Rate}
\label{sec:masking}
As was done in \cite{maskgit}, we train our model with a variable masking rate based on a Cosine scheduling: for each training example, we sample a masking rate $r\in[0,1]$ from a truncated $\arccos$ distribution with density function $p(r)=\frac{2}{\pi}(1-r^2)^{-\frac{1}{2}}$. This has an expected masking rate of 0.64, with a strong bias towards higher masking rates. The bias towards higher masking rates makes the prediction problem harder. In contrast with autoregressive approaches, which learn conditional distributions $P(x_i | x_{<i})$ for some fixed ordering of tokens, random masking with a variable masking ratio allows our models to learn $P(x_i | x_{\Lambda})$ for arbitrary subsets of tokens $\Lambda$. This is not only critical for our parallel sampling scheme, but it also enables a number of zero-shot, out-of-the-box editing capabilities, such as shown in \figg{teaser_edit} and \secc{editing}.

\subsection{Classifier Free Guidance}
\label{sec:cfg}
We employ classifier-free guidance (CFG) \citep{ho2022classifier} to improve our generation quality and our text-image alignment. At training time, we remove text conditioning on 10\% of samples chosen randomly (thus attention reduces to image token self-attention). At inference time, we compute a conditional logit $\ell_c$ and an unconditional logit $\ell_u$ for each masked token. We then form the final logits $\ell_g$ by moving away from the unconditional logits by an amount $t$, the \emph{guidance scale}:
\begin{equation}
    \ell_g = (1+t)\ell_c - t \ell_u
    \label{eq:cfg}
\end{equation}
Intuitively, CFG trades off diversity for fidelity. Different from previous approaches, we reduce the hit to diversity by linearly increasing the guidance scale $t$ through the sampling procedure. This allows the early tokens to be sampled more freely, with low or no guidance, but increases the influence of the conditioning prompt for the later tokens.

We also exploit this mechanism to enable \emph{negative prompting} \citep{negprompt} by replacing the unconditional logit $\ell_u$ with a logit conditioned on a ``negative prompt''. This encourages the resulting image to have features associated with the positive prompt $\ell_c$ and remove features associated with the negative prompt $\ell_u$. 
\subsection{Iterative Parallel Decoding at Inference}
\label{sec:iterativedec}
\begin{figure*}[ht!]
\begin{center}
\vspace{-5pt}
\includegraphics[width=1.0\textwidth]{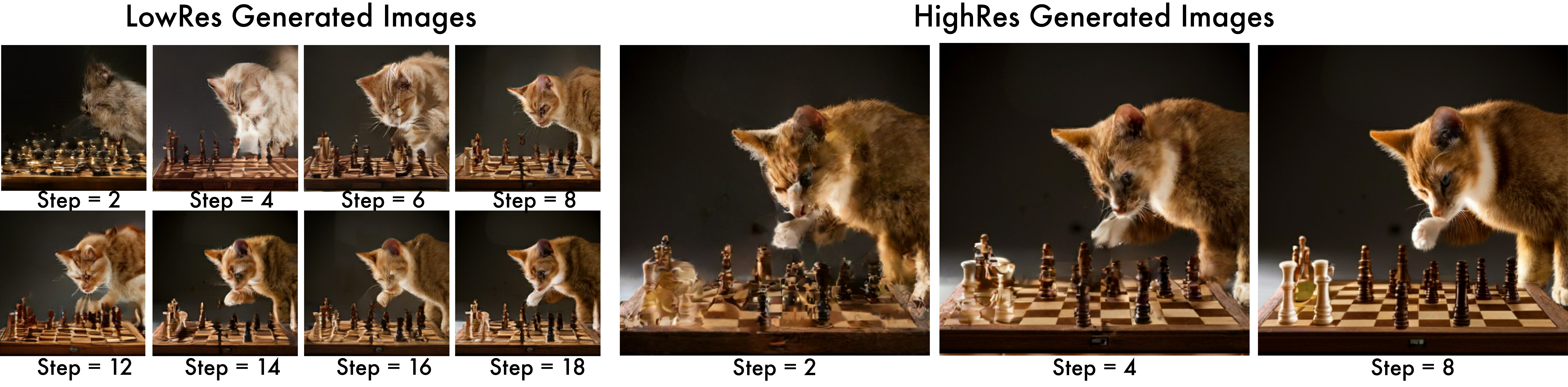}
\end{center}
\vspace{-10pt}
\caption{\small Inference samples. We visualize the evolution of masked tokens over the sequence of steps for the base model (left) and the super-res model (right). The super-res model, being conditioned on the low-res tokens, requires significantly fewer sampling steps for convergence. 
}
\vspace{-5pt}
\label{fig:srmodel}
\end{figure*}
The critical component for our model's inference time efficiency is the use of parallel decoding to predict multiple output tokens in a single forward pass. The key assumption underlying the effectiveness of the parallel decoding is a Markovian property that many tokens are conditionally independent given other tokens. Decoding is performed based on a cosine schedule \citep{maskgit} that chooses a certain fixed fraction of the highest confidence masked tokens that are to be predicted at that step. These tokens are then set to unmasked for the remainder of the steps and the set of masked tokens is appropriately reduced. Using this procedure, we are able to perform inference of $256$ tokens using only $24$ decoding steps in our base model and $4096$ tokens using $8$ decoding steps in our super-resolution model, as compared to the 256 or 4096 steps required for autoregressive models (e.g. \citep{parti}) and hundreds of steps for diffusion models (e.g., \citep{ldm,imagen}). We note that recent methods including progressive distillation \citep{salimans2022distillation} and better ODE solvers \citep{Zhu2022dpm} have greatly reduced the sampling steps of diffusion models, but they have not been widely validated in large scale text-to-image generation. We leave the comparison to these faster methods in the future work, while noting that similar distillation approaches are also a possibility for our model.

\section{Results}
\label{sec:results}
We train a number of base Transformer models at different parameter sizes, ranging from 600M to 3B parameters. Each of these models is fed in the output embeddings from a T5-XXL model, which is pre-trained and frozen and consists of 4.6B parameters. Our largest base model of 3B parameters consists of $48$ Transformer layers with cross-attention from text to image and self-attention among image tokens. All base models share the same image tokenizer. We use a CNN model with $19$ ResNet blocks and a quantized codebook of size $8192$ for the tokenization. Larger codebook sizes did not result in performance improvements. 
The super-resolution model consists of $32$ multi-axis Transformer layers~\cite{hit} with cross-attention from concatenated text and image embedding to high resolution image and self-attention among high resolution image tokens. This model converts a sequence of tokens from one latent space to another: the first latent space being that of the base model tokenizer, a latent space of $16\times16$ tokens, to that of a higher resolution tokenizer with $64\times64$ tokens. After token conversion, the decoder for the higher resolution tokenizer is used to convert to the higher resolution image space. Further details of configurations are provided in the appendix.

We train on the Imagen dataset consisting of 460M text-image pairs \citep{imagen}.
Training is performed for 1M steps, with a batch size of 512 on 512-core TPU-v4 chips \citep{jouppi2020domain}. This takes about 1 week of training time. We use the Adafactor optimizer \citep{shazeer2018adafactor} to save on memory consumption which allowed us to fit a 3B parameter model without model parallelization. We also avoid performing exponential moving averaging (EMA) of model weights during training, again to save on TPU memory. In order to reap the benefits of EMA, we checkpoint every 5000 steps, then perform EMA offline on the checkpointed weights with a decay factor of 0.7. These averaged weights form the final base model weights.

\begin{figure}
  \centering
  \begin{tabular}{p{25mm}p{25mm}p{25mm}|p{25mm}p{25mm}p{25mm}}
    \multicolumn{3}{c}{Cardinality} &
    \multicolumn{3}{c}{Composition}
    \\
    \includegraphics[width=26mm]{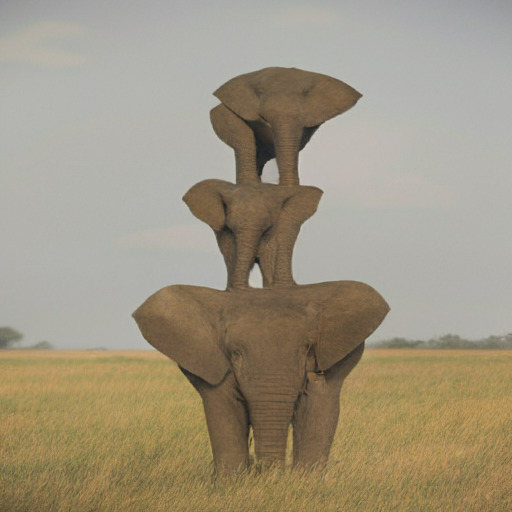} &       \includegraphics[width=26mm]{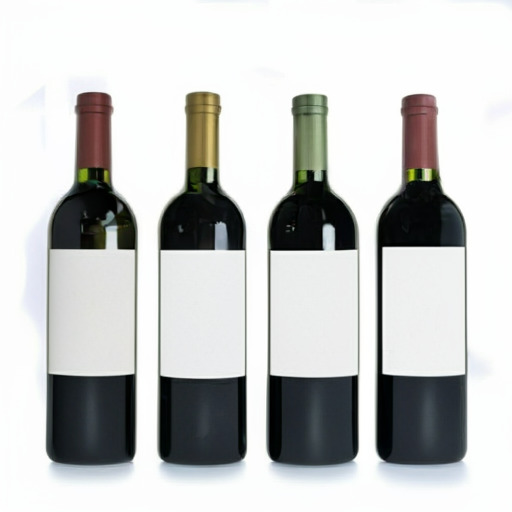} &
    \includegraphics[width=26mm]{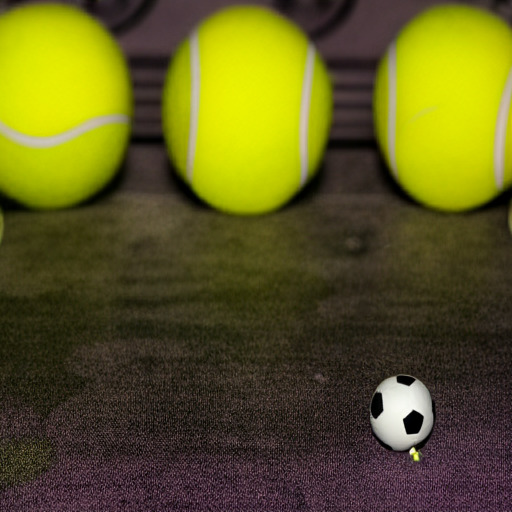} &
    \includegraphics[width=26mm]{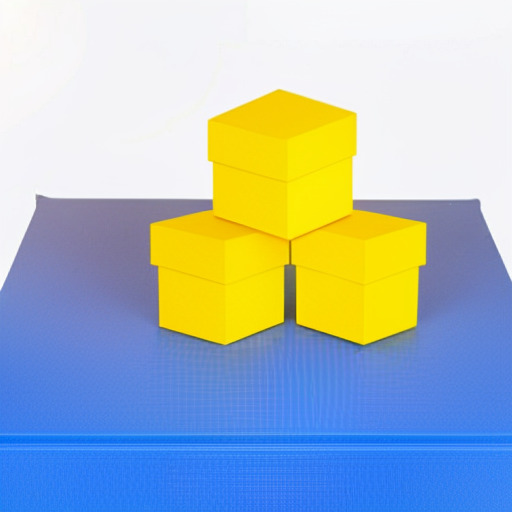} &
    \includegraphics[width=26mm]{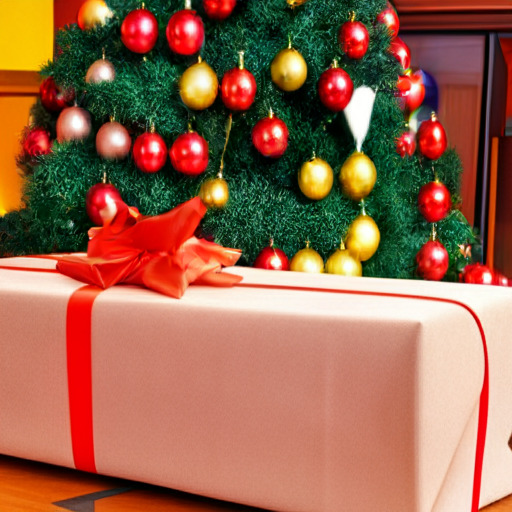} &
    \includegraphics[width=26mm]{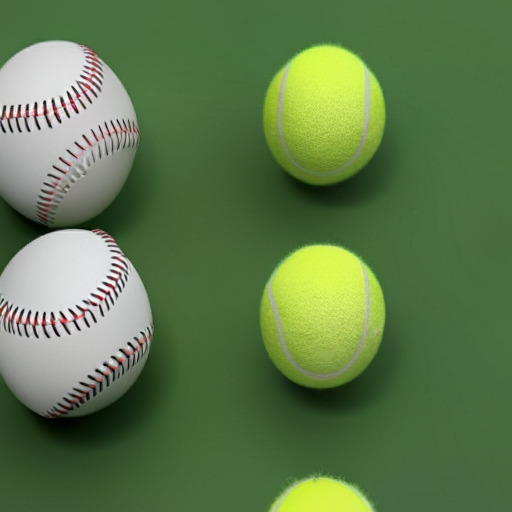}
    \\
    \tiny Three elephants standing on top of each other. &
    \tiny Four wine bottles. &
    \tiny A tiny football in front of three yellow tennis balls. &
    \tiny Three small yellow boxes on a large blue box. &
    \tiny A large present with a red ribbon to the left of a Christmas tree. &
    \tiny Two baseballs to the left of three tennis balls.
    \\
    \noalign{\vskip 2mm}
    \multicolumn{3}{c}{Style} &
    \multicolumn{3}{c}{Text Rendering}
    \\
    \includegraphics[width=26mm]{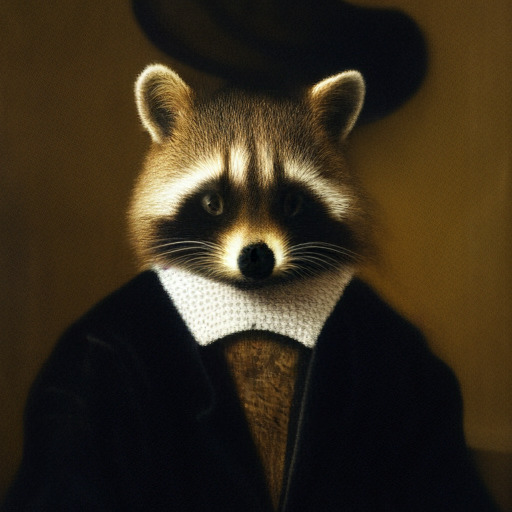} &
    \includegraphics[width=26mm]{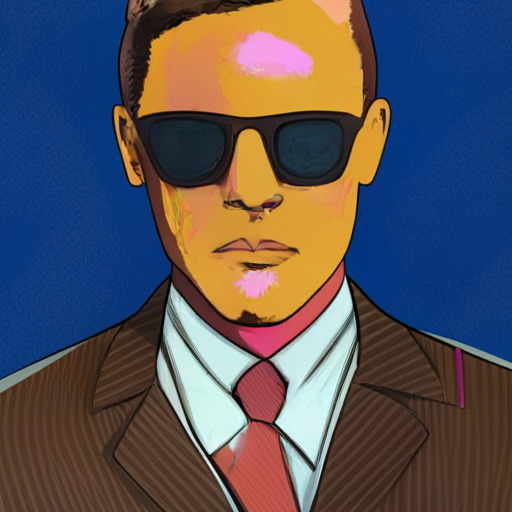} &
    \includegraphics[width=26mm]{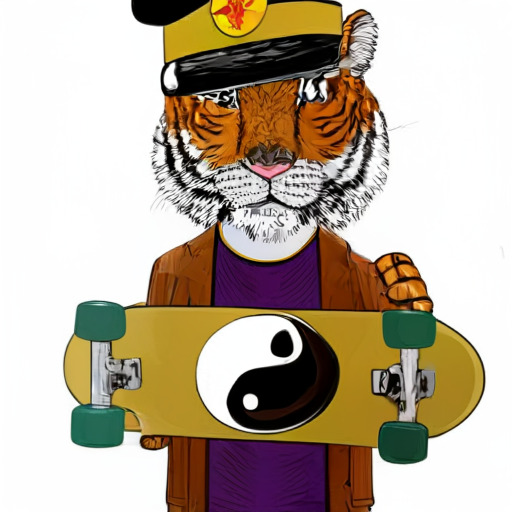} &
    \includegraphics[width=26mm]{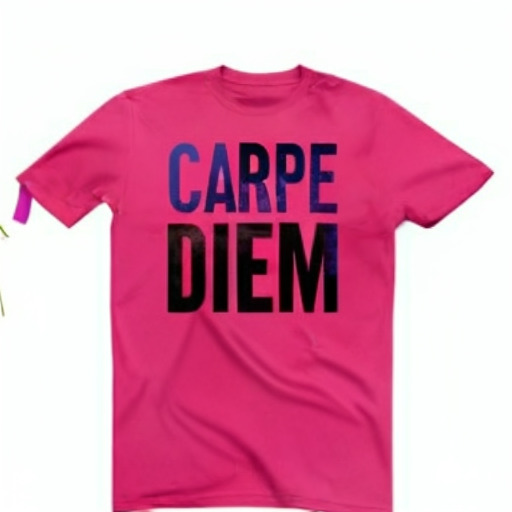} &
    \includegraphics[width=26mm]{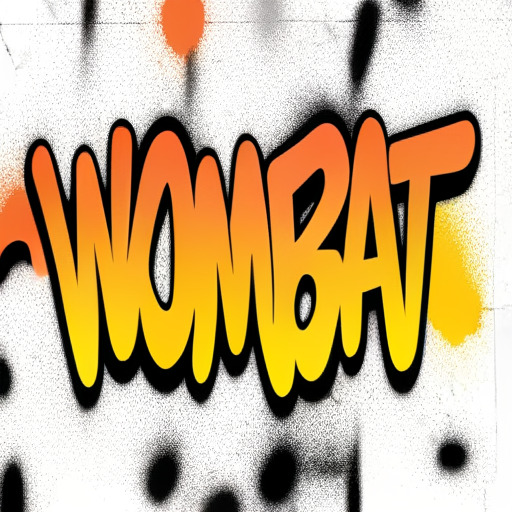} &
    \includegraphics[width=26mm]{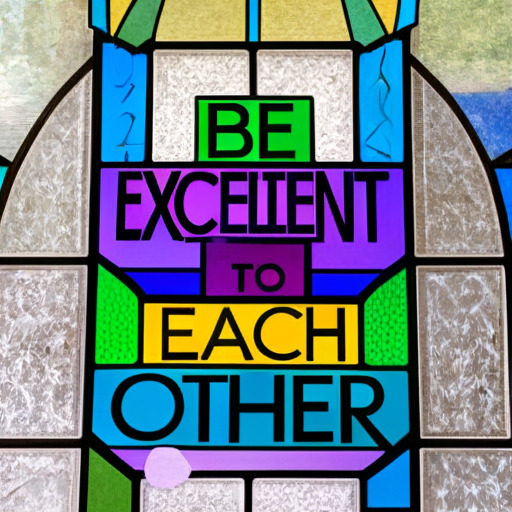}
    \\
    \tiny Portrait of a well-dressed raccoon, oil painting in the style of Rembrandt. &
    \tiny A portrait of a man wearing sunglasses and a business suit, painting in pop art style. &
    \tiny Portrait of a tiger wearing a train conductor’s hat and holding a skateboard that has a yin-yang symbol on it. Chinese ink and wash painting. &
    \tiny A t-shirt with Carpe Diem written on it. &
    \tiny High-contrast image of the word ``WOMBAT'' written with thick colored graffiti letters on a white wall with dramatic splashes of paint. &
    \tiny The saying ``BE EXCELLENT TO EACH OTHER'' written in a stained glass window.
    \\
    \noalign{\vskip 2mm}
    \multicolumn{3}{c}{Usage of Entire Prompt} &
    \multicolumn{3}{c}{Failure Text Classes}
    \\
    \includegraphics[width=26mm]{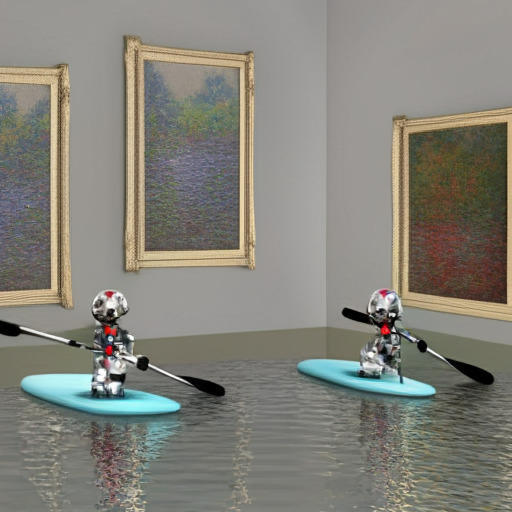} &
    \includegraphics[width=26mm]{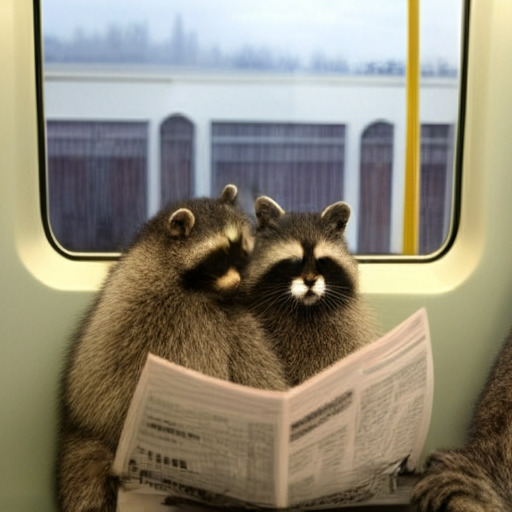} &  
    \includegraphics[width=26mm]{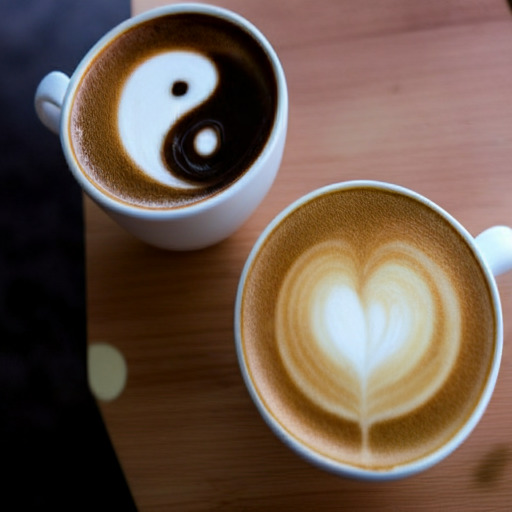} &
    \includegraphics[width=26mm]{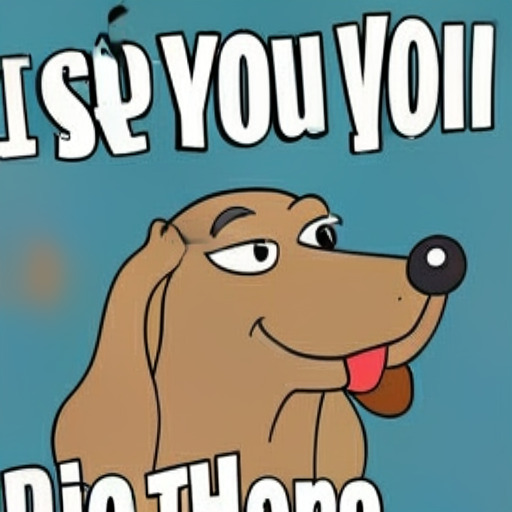} &
    \includegraphics[width=26mm]{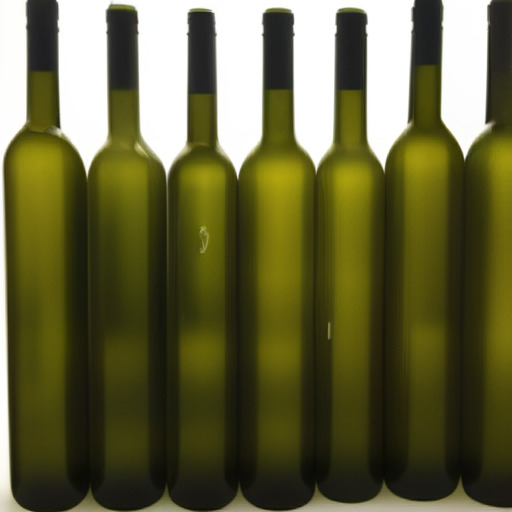} &
    \includegraphics[width=26mm]{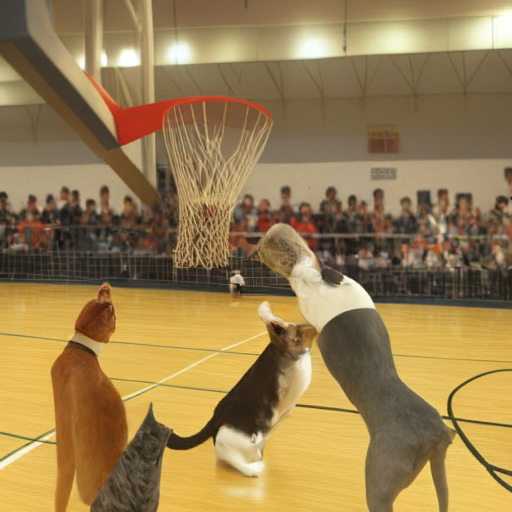}
    \\
    \tiny An art gallery displaying Monet paintings. The art gallery is flooded. Robots are going around the art gallery using paddle boards. &
    \tiny A photograph of the inside of a subway train. There are raccoons sitting on the seats. One of them is reading a newspaper. The window shows the city in the background. &
    \tiny Two cups of coffee, one with latte art of yin yang symbol. The other has latter art of a heart. &
    \tiny A cartoon of a dog saying ``I see what you did there''. &
    \tiny Ten wine bottles. &
    \tiny A basketball game between a team of four cats and a team of three dogs.
    \\
  \end{tabular}
  \caption{\small Examples demonstrating text-to-image capabilities of \name~for various text properties. Top left: cardinality; top right: composition; middle left: style; middle right: text rendering; and bottom left: usage of the entire prompt. For all examples, $16$ instances per prompt were generated, and the one with the highest CLIP score \citep{clip} was chosen. Bottom right: examples of generated image failure in \name~for various text properties such as direct rendering of long phrases, high cardinalities, and multiple cardinalities.}
  \label{fig:text_class_examples}
\end{figure}

\renewcommand{\figwidth}{0.93\textwidth}
\begin{figure*}[htbp!]
\vspace{-10pt}
\centering
\captionsetup{width=\figwidth}
\includegraphics[width=\figwidth]{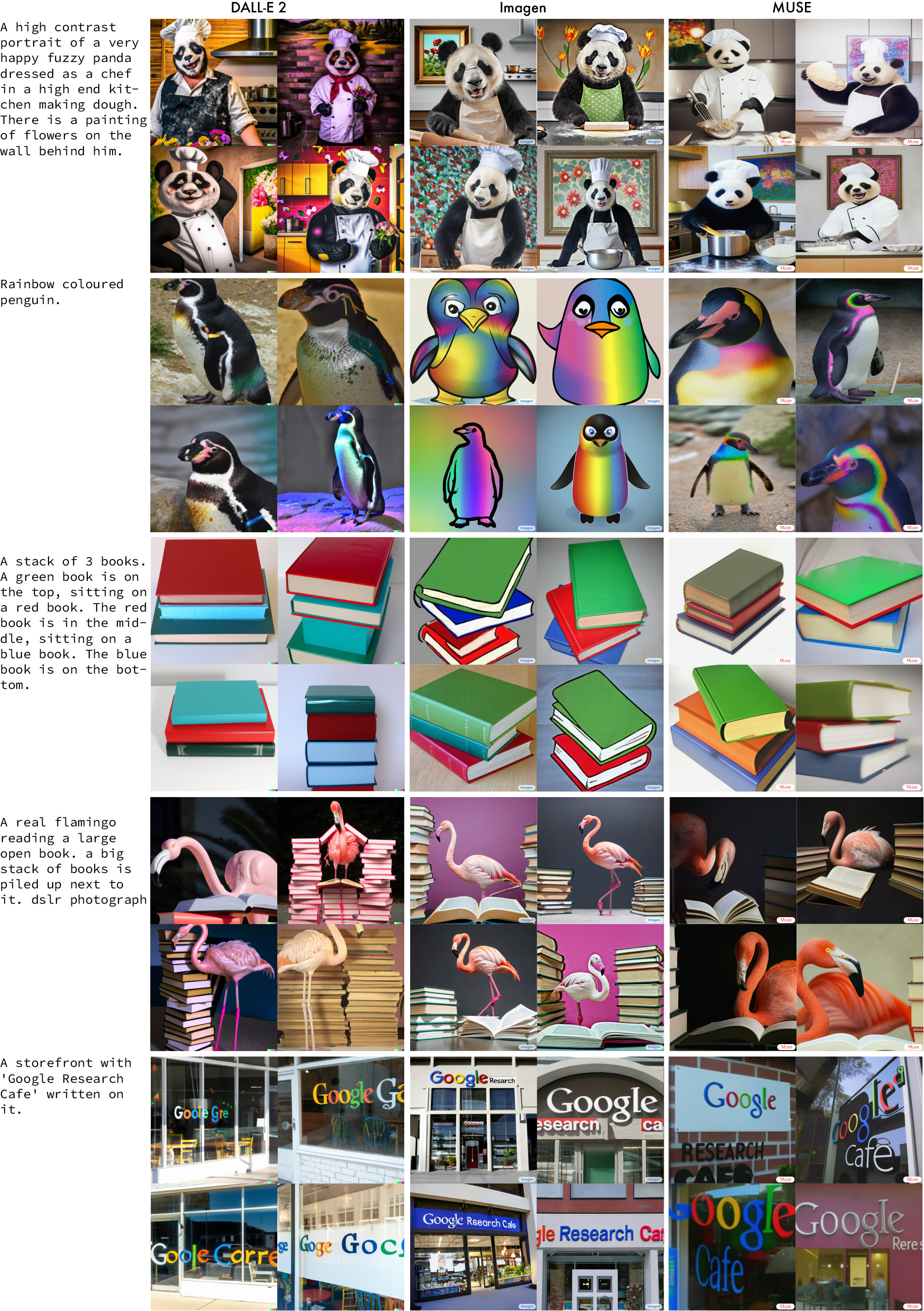}
\vspace{-5pt}
\caption{\small Comparing the same prompts across DALL-E2 \citep{dalle2} (left), Imagen \citep{imagen} (middle) and \name~(right).
}
\vspace{-10pt}
\label{fig:comparison}
\end{figure*}

\subsection{Qualitative Performance}

\figg{text_class_examples} qualitatively demonstrates the capabilities of \name~for text prompts with different properties. The top left of \cref{fig:text_class_examples} shows examples that demonstrate a basic understanding of cardinality. For objects with non-unity cardinality, instead of generating the same object pixels multiple times, \name~instead adds contextual variations to make the overall image more realistic, e.g., elephant size and orientation, wine bottle wrapper color, and tennis ball rotation. The top right of Fig, \ref{fig:text_class_examples} demonstrates understanding of multi-object composition and relativeness. Instead of placing objects at random locations, \name~generates images that preserve prepositional object relations in the text, e.g., on vs under, left vs right, etc. The middle left of \cref{fig:text_class_examples} demonstrates its ability to generate images spanning many styles, both specific to a renowned artist (e.g., Rembrandt) as well as general to a style as a whole (e.g., pop art and Chinese ink and wash). The middle right of \cref{fig:text_class_examples} demonstrates the ability of \name~to render words and phrases. Text generation is fundamentally different than generating most other objects. Instead of the model learning a mapping between an object name and its characteristics (e.g., that ``elephant'' maps to ``large'', ``gray'', and ``peanut eating''), the virtual continuum of possible words and phrases demands that the model learn differently. It must instead learn a hierarchical understanding between phrases, words, and letters. The bottom left of \cref{fig:text_class_examples} demonstrates that \name~uses the entirety of a text prompt when rendering instead of focusing exclusively on only a few salient words. Finally, \cref{fig:comparison} shows comparisons between \name, Dall-E 2 \citep{dalle2}, and Imagen \citep{imagen} for some select prompts, showing that \name~is at par with Imagen and qualitatively better than Dall-E2 for many prompts.

However, as demonstrated in the bottom right of \cref{fig:text_class_examples}, \name~is limited in its ability to generate images well aligned with certain types of prompts. For prompts which indicate that long, multi-word phrases should be directly rendered, ~\name~has a tendency to render those phrases incorrectly, often resulting in (unwanted) duplicated rendered words or rendering of only a portion of the phrase. Additionally, prompts indicating high object cardinality tend to result in generated images which do not correctly reflect that desired cardinality (e.g., rendering only $7$ wine bottles when the prompt specified $10$). In general, the ability of \name~to render the correct cardinalities of objects decreases as the cardinality increases. Another difficult prompt type for \name~is ones with multiple cardinalities (e.g., ``four cats and a team of three dogs''). For such cases, \name~has a tendency to get at least one cardinality incorrect in its rendering.

\subsection{Quantitative Performance}

\begin{table}[t]

\vspace{5pt}
\label{tab:cc3m}
\begin{center}{
  \begin{tabular}{p{50mm} |c | r | r r }
\toprule
 \bfseries{Approach} &
 \bfseries{Model Type} &
 \bfseries{Params} & 
 \bfseries{FID} & 
 \bfseries{CLIP} \\ 
\toprule
VQGAN~\cite{esser2021taming} & Autoregressive &  600M & 28.86 & 0.20 \\
ImageBART~\citep{esser2021imagebart} & Diffusion+Autogressive & 2.8B & 22.61 & 0.23 \\
LDM-4~\citep{ldm} & Diffusion &645M & 17.01 & 0.24 \\
RQ-Transformer \citep{lee2022autoregressive} & Autoregressive & 654M & 12.33 & 0.26 \\
Draft-and-revise \citep{lee2022draft} & Non-autoregressive  & 654M & 9.65 & 0.26 \\
\midrule
\textbf{\name (base model)} & Non-autoregressive & 632M & 6.8 & 0.25 \\
\textbf{\name (base + super-res)} & Non-autoregressive & 632M + 268M & 6.06 & 0.26 \\
\bottomrule
\end{tabular}
}
\end{center}
\caption{\small Quantitative evaluation on CC3M \citep{sharma2018conceptual}; all models are trained and evaluated on CC3M.}
\label{tab:eval_cc3m}
\vspace{-10pt}
\end{table}


\begin{table}[t]

\vspace{5pt}
\centering
\label{tab:zero_shot_mscoco}
\begin{tabular}{p{50mm}|c| r | r r}
\toprule
\multirow{2}{*}{\bfseries{Approach}} & \multirow{2}{*}{\bfseries{Model Type}} &
\multirow{2}{*}{\bfseries{Params}} &
\multirow{2}{*}{\bfseries{FID-30K}} &
\bfseries{Zero-shot} \\
& & & & \bfseries{FID-30K} \\
\midrule
AttnGAN \citep{attngan} & GAN & & 35.49 & - \\
DM-GAN \citep{zhu2019dm} & GAN & & 32.64 & - \\
DF-GAN \citep{dfgan} &  GAN & & 21.42 & - \\
DM-GAN + CL \citep{dmgan-cl} &  GAN & & 20.79 & - \\
XMC-GAN \citep{zhang2021cross} & GAN & & 9.33 & - \\
LAFITE \citep{lafite} & GAN & & 8.12 & - \\
Make-A-Scene \citep{makeascene} & Autoregressive & & 7.55 & -\\
\midrule
DALL-E \citep{dalle1} & Autoregressive & & - & 17.89  \\
LAFITE \citep{lafite} & GAN & & - & 26.94 \\
LDM \citep{ldm}  & Diffusion & & -  & 12.63 \\
GLIDE \citep{glide} & Diffusion & & -  & 12.24 \\
DALL-E 2 \citep{dalle2} & Diffusion & & -  & 10.39 \\
Imagen-3.4B \citep{imagen}  & Diffusion & & -  & 7.27 \\ 

Parti-3B \citep{parti} & Autoregressive & & -  & 8.10 \\
Parti-20B \citep{parti} & Autoregressive & & 3.22  & 7.23\\
\midrule
\textbf{\name-3B} & Non-Autoregressive & & - & \cocofid \\
\bottomrule
\end{tabular}

\caption{\small Quantitative evaluation of FID and CLIP score (where available) on MS-COCO \citep{coco} for $256\times256$ image resolution. \name~ achieves a CLIP score of 0.32, higher than the score of 0.27 reported in Imagen. Other papers in the table above did not report a CLIP score.}
\label{tab:eval_coco}
\end{table}

\begin{figure}
  \centering
  \begin{tabular}{p{0.4\textwidth}p{0.4\textwidth}}
    \includegraphics[width=0.4\textwidth,height=0.3\textwidth]{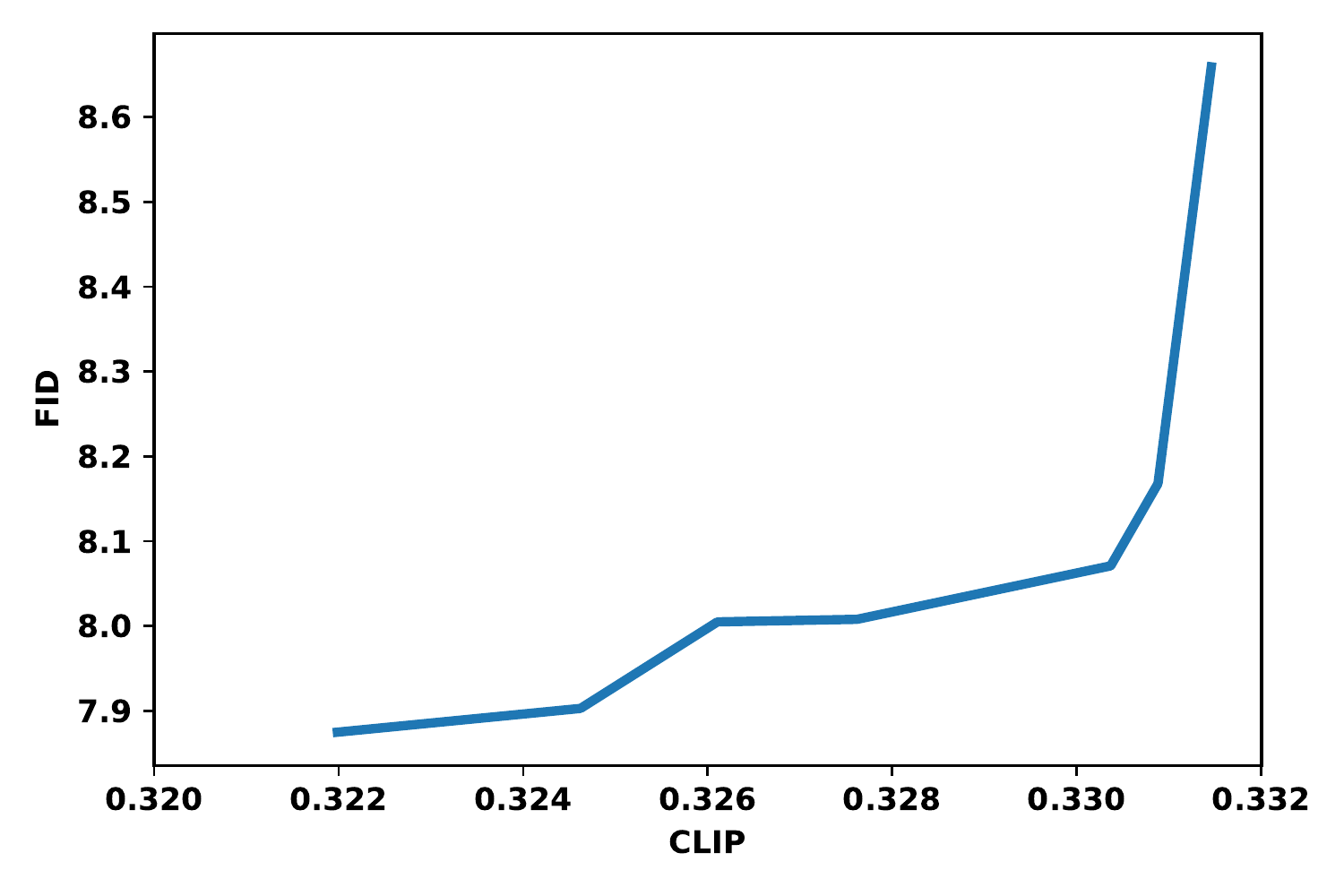}
    \caption{CLIP vs. FID tradeoff curve. We perform sweeps of sampling parameters for a fixed model, then plot the Pareto front.}
    \label{fig:pareto_curve}
    &
    \includegraphics[width=0.4\textwidth,height=0.3\textwidth]{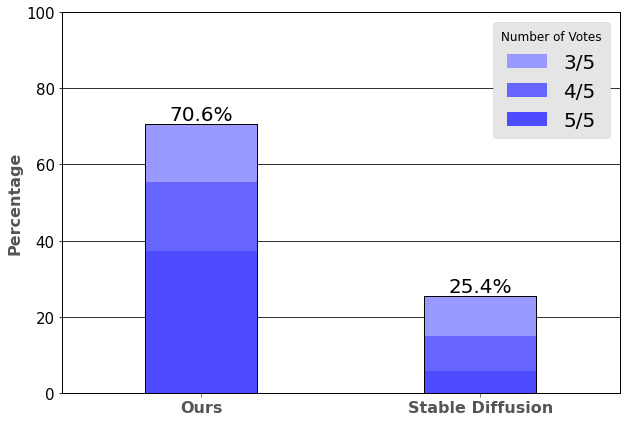}
    \caption{\small Percentage of prompts for which a human rater consensus chose a model alignment preference. Contributions from specific numbers of rater consensuses are shown in different colors, while marginals over consensuses ($=\!5$, $\geq4$, and $\geq3$) are shown numerically.}
    \label{fig:rater_percentages}
  \end{tabular}
\end{figure}

In \tabb{eval_cc3m} and \tabb{eval_coco}, we show our performance against other methods on the CC3M \citep{sharma2018conceptual} and COCO \citep{coco} datasets as measured by Fréchet Inception Distance (FID) \citep{fid}, which measures quality and diversity of samples, as well as CLIP \citep{clip} score, which measures image/text alignment. For the CC3M results, both \name~ models were trained on CC3M. The COCO results are zero-shot, using a model trained on the same dataset as Imagen \citep{imagen}. 

Our 632M model achieves SOTA results on CC3M, significantly improving upon the state of the art in FID score, and also achieving state of the art CLIP score. Our 3B model achieves an FID score of \cocofid~which is slightly better than the score of $8.1$ achieved by the Parti-3B model which has a similar number of parameters. Our CLIP score of \cococlip~is higher than the CLIP score of 0.29 achieved by Imagen (which is achieved when the FID is significantly higher ~20). For the FID of 7.27, Imagen achieves a CLIP score of around 0.27 (see Figure 4 in \citep{imagen}).

Our sampling algorithm (\cref{sec:iterativedec}) has a number of hyperparameters, such as guidance scale, sampling temperature, whether or not to linearly increase guidance during sampling, etc. We perform evaluation sweeps over these parameters. We find subsets of sampling parameters that are Pareto efficient, in the sense that we cannot improve FID without hurting CLIP. This allows us to study the tradeoff between diversity and image/text alignment, which we show in \cref{fig:pareto_curve}.

\subsubsection{Human evaluation}

Similar to previous works \citep{parti, imagen}, we perform side-by-side evaluations in which human raters are presented with a text prompt and two images, each generated by a different text-to-image model using that prompt. The raters are asked to assess prompt-image alignment via the question, ``Which image matches with the caption better?'' Each image pair is anonymized and randomly ordered (left vs right). Raters have the option of choosing either image or that they are indifferent\footnote{Choosing indifference makes sense when neither image is aligned with the text prompt and helps reduce statistical noise in the results.}. Each (prompt, image pair) triplet is assessed by five independent raters; the raters were provided through the Google internal crowd computing team and were completely anonymous to the \name~team. For the set of prompts presented to raters, we used PartiPrompts \citep{parti}, a collection of $1650$ text prompts curated to measure model capabilities across a variety of categories. For the two text-to-image models, we compared \name~($3$B parameters) to that of Stable Diffusion v1.4 \citep{ldm}, the text-to-image model most comparable to \name~in terms of inference speed. For each prompt, $16$ image instances were generated, and the one with the highest CLIP score \citep{clip} was used. The stable diffusion images were generated via the CompVis Stable Diffusion v1.4 notebook \citep{sdgeneration}. We required at least a $3$ rater consensus for results to be counted in favor of a particular model. From this analysis, we found that \name~was chosen as better aligned than Stable Diffusion for $70.6$\% of the prompts, Stable Diffusion was chosen as better aligned than \name~for $25.4$\%, and no rater consensus was chosen for $4$\%. These results are consistent with \name~having significantly better caption matching capability ($\sim\!2.7$x). \cref{fig:rater_percentages} shows a breakdown of the rater results for rater consensuses of $3$, $4$, and all $5$ possible votes. Prompts for which all $5$ raters said \name~had better alignment than Stable Diffusion are the larger contributor.

In addition to measuring alignment, other works \citep{parti, imagen} have also measured image realism, often via a rater question similar to, ``Which image is more realistic?''. However, we note that care must be taken with examination of such results. Though it is not the intent of the question, a model that is completely mode collapsed so that it generates the same sufficiently realistic image regardless of prompt will virtually always do better on this question than a model that \textit{does} take the prompt into account during image generation. We propose this type of question is only applicable between models of similar alignment. Since \name~is significantly better aligned than Stable Diffusion, we did not assess realism via human raters. We consider this topic an area of open research.
\newcommand{\pz}{\hphantom{0}}
\begin{wraptable}{R}{0.5\textwidth}
    \vspace{-30pt}
    \centering
    \begin{tabular}{c|c|r}
         \textbf{Model} & \textbf{Resolution} & \textbf{Time}  \\
         \hline
         Imagen & \lowressq &  9.1s \\
         Imagen & $1024\times 1024$ &  13.3s \\
         LDM (50 steps) & $512\times 512$ & 3.7s \\
         LDM (250 steps) & $512\times 512$ & 18.5s \\
         Parti-3B& $256\times256$ & 6.4s \\
         \hline
         \name-3B& \lowressq & 0.5s \\
         \name-3B& \highressq & 1.3s \\
    \end{tabular}
    \vspace{-5pt}
    \caption{\small Per-batch inference time for several models. Muse, Imagen, and Parti were benchmarked internally on TPUv4 hardware. Stable Diffusion/LDM benchmark from \cite{sdinference}, on A100 GPUs. The ``LDM (250 steps)'' time comes from scaling the 50-step time by 5; 250 steps were used to achieve the FID in \cref{tab:eval_coco}.}
    \label{tbl:speed}
\end{wraptable}

\subsubsection{Inference Speed}
\label{sec:speed}
In \cref{tbl:speed}, we compare the inference time of \name~to several other popular models. We benchmarked Parti-3B, Imagen, and Muse-3B internally on TPUv4 accelerators. 
For Stable Diffusion/LDM, we used the fastest reported benchmark \cite{sdinference}, which was done on A100 GPUs. For Stable Diffusion, the TPU implementations we tested were not faster than the A100 implementation. We also report an inference time for LDM with 250 iterations, which is the configuration used to achieve the FID in \cref{tab:eval_coco}. \name~is significantly faster than competing diffusion or autoregressive models, despite having comparable parameter counts (and around 3x more parameters than Stable Diffusion/LDM). The speed advantage of \name~over Imagen is due to the use of discrete tokens and requiring fewer sampling iterations. The speed advantage of \name~over Parti is due to the use of parallel decoding. The speed advantage of \name~over Stable Diffusion is primarily attributable to requiring fewer sampling iterations. 
\subsection{Image Editing}
\label{sec:editing}
By exploiting the fact that our model can condition on arbitrary subsets of image tokens, we can use the model out-of-the-box for a variety of image editing applications with no additional training or model fine-tuning.
\begin{figure*}
  \centering
  \begin{tabular}{p{27mm}p{27mm}p{27mm}p{27mm}p{27mm}}
    \includegraphics[width=30mm]{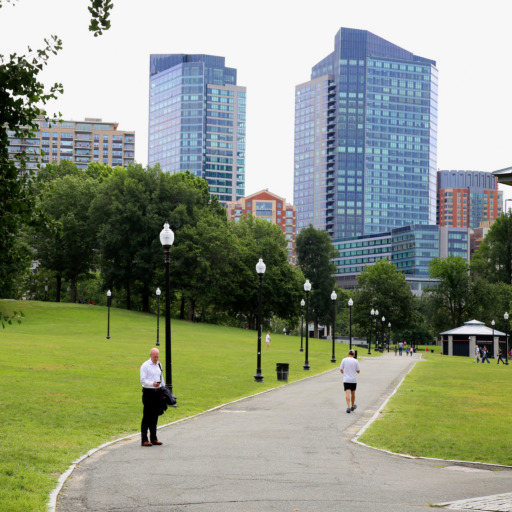} &
    \includegraphics[width=30mm]{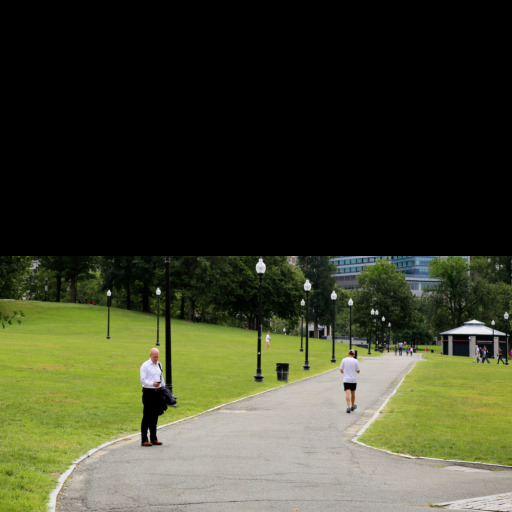} &
    \includegraphics[width=30mm]{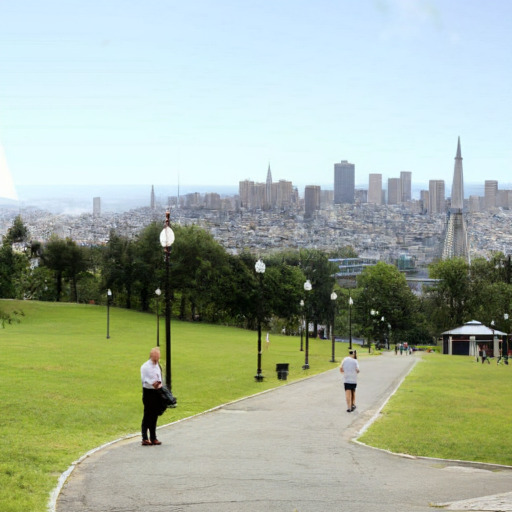} &
    \includegraphics[width=30mm]{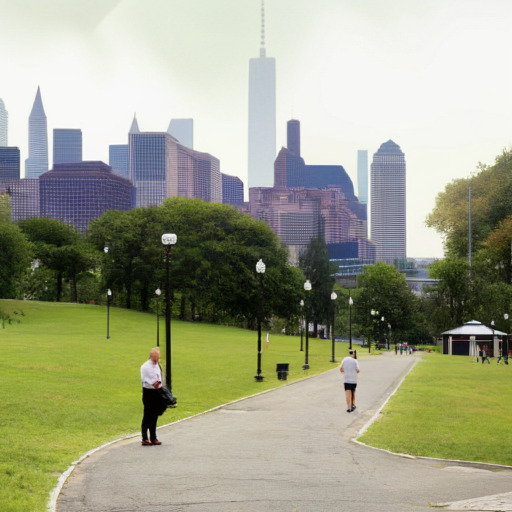} &
    \includegraphics[width=30mm]{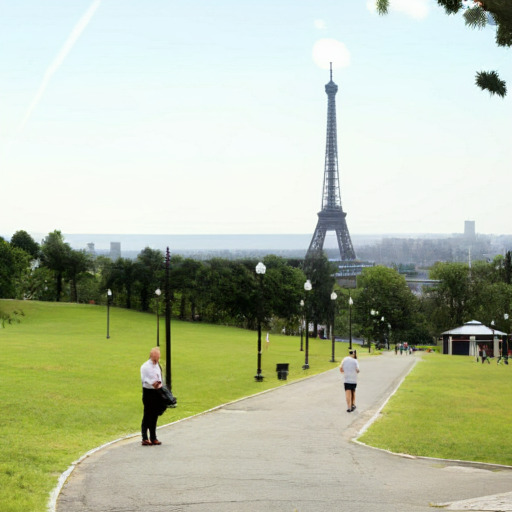} 
    \\
     Original &
     Masked &
    {San Francisco in the background} & 
    {New York City in the background} &
    {Paris in the background} 
    \\
    \includegraphics[width=30mm]{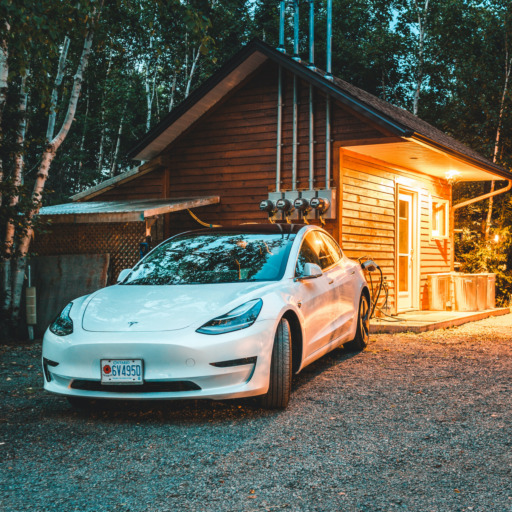} &
    \includegraphics[width=30mm]{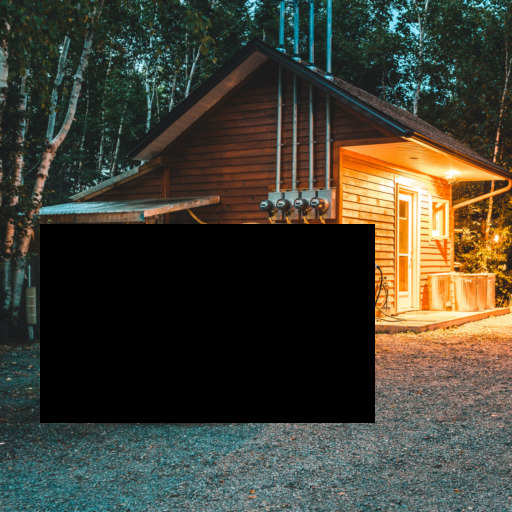} &
    \includegraphics[width=30mm]{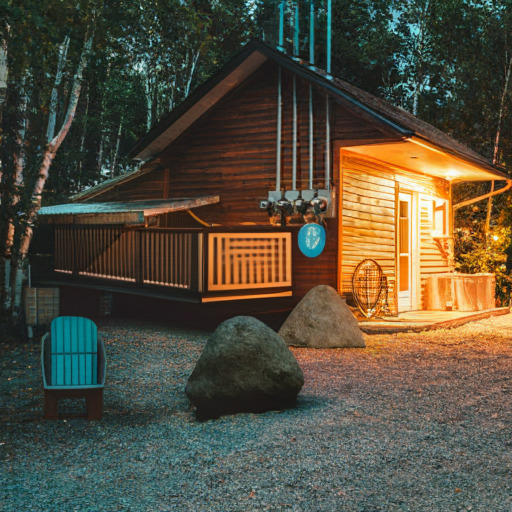} &
    \includegraphics[width=30mm]{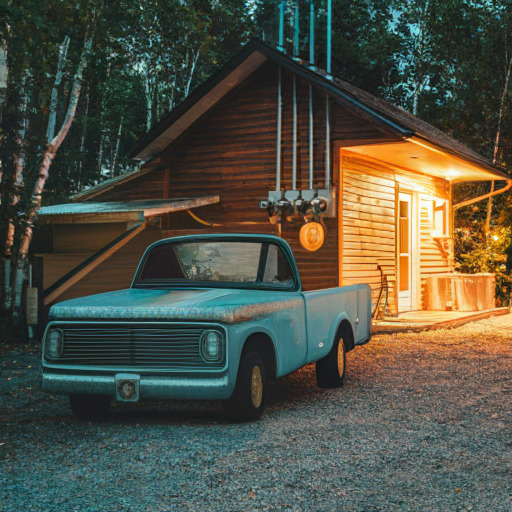} &
    \includegraphics[width=30mm]{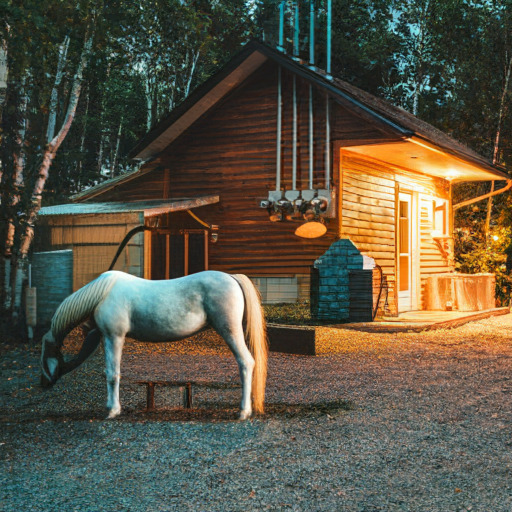} 
    \\
    Original &
    Masked &
    {A cabin in the woods} & 
    {An old, beat up pickup truck.} & 
    {A horse tied to a post.} 
  \end{tabular}
  \caption{\small Examples of text-guided inpainting. The mask is shown in the second column of each row. This behavior arises directly from the model with no fine-tuning.}
  \label{fig:inpainting}
\end{figure*}
\subsubsection{Text-guided Inpainting / outpainting}
Our sampling procedure (\cref{sec:iterativedec}) gives us text-guided inpainting and outpainting for free: we convert an input image into a set of tokens, mask out the tokens corresponding to a local region, and then sample the masked tokens conditioned on unmasked tokens and a text prompt. We integrate superresolution through a multi-scale approach: Given an image of size 512x512, we first decimate it to 256x256 and convert both images to high- and low-res tokens. Then, we mask out the appropriate regions for each set of tokens. Next, we inpaint the low-res tokens using the parallel sampling algorithm. Finally, we condition on these low-res tokens to inpaint the high-res tokens using the same sampling algorithm. We show examples of this in \cref{fig:teaser_edit} and \cref{fig:inpainting}.

\subsubsection{Zero-shot Mask-free editing}
We use \name~in a zero-shot sense for mask-free image editing of real input images. This method works directly on the (tokenized) image and does not require ``inverting'' the full generative process, in contrast with recent zero-shot image editing techniques leveraging generative models \citep{gal2022stylegan,patashnik2021styleclip,kim2022diffusionclip,nulltext2022}. 

We first convert an input image into visual tokens. Next, we iteratively mask and resample a random subset of tokens, conditioned on text prompts. We can think of this as being analogous to a Gibbs sampling procedure, where we fix some tokens and resample others conditioned on them. This has the effect of moving the tokenized image into the typical set of the conditional distribution of images given a text prompt. 

\begin{figure*}
  \centering
  \begin{tabularx}{0.95\textwidth}{p{0mm}p{27mm}p{27mm}p{27mm}p{27mm}p{27mm}}
    \rotatebox{90}{\hspace{5mm}Input image} &
    \includegraphics[width=30mm]{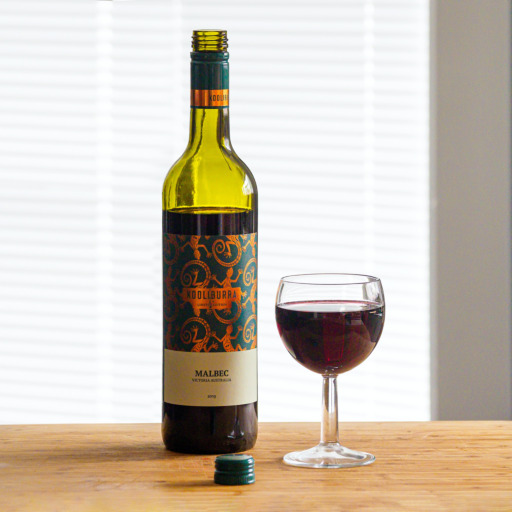} &
    \includegraphics[width=30mm]{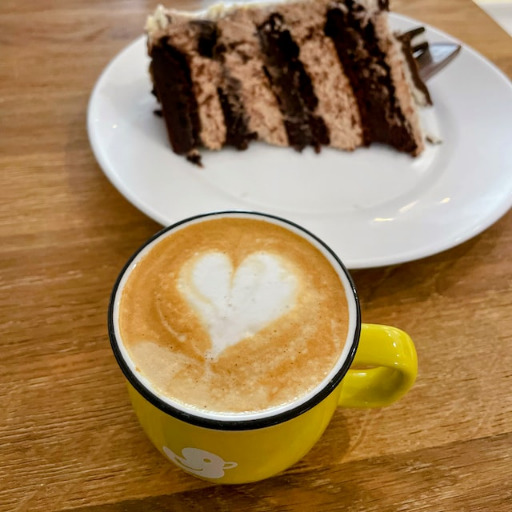} &
    \includegraphics[width=30mm]{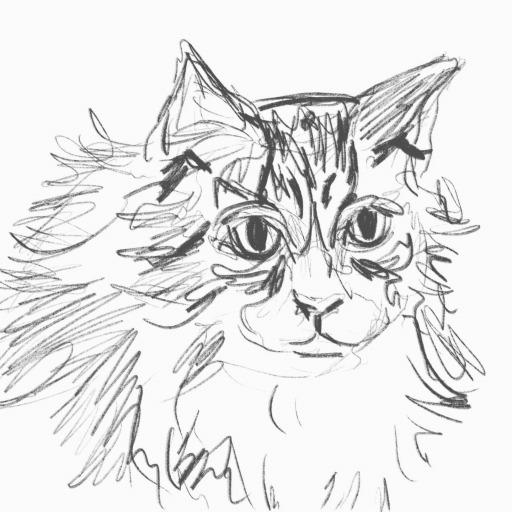} &
    \includegraphics[width=30mm]{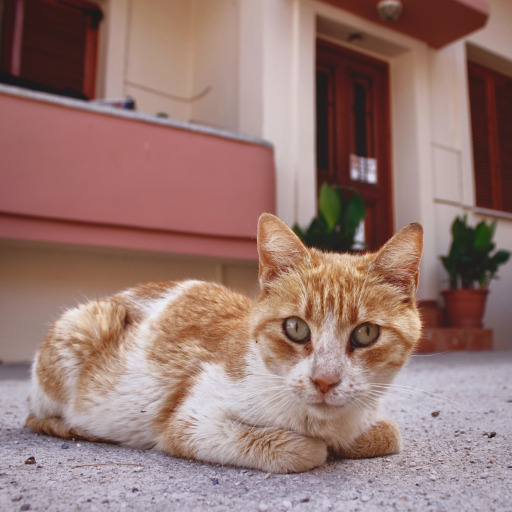} &
    \includegraphics[width=30mm]{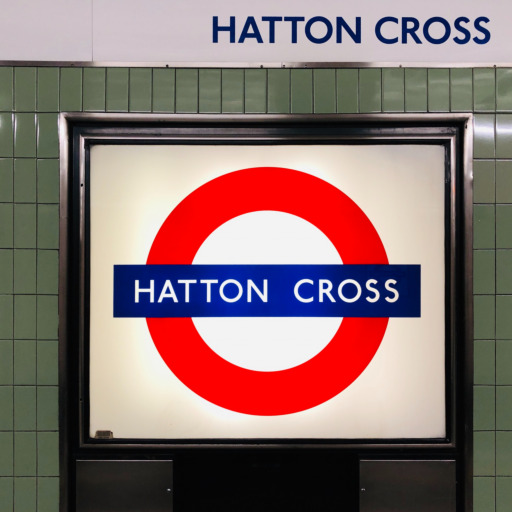} 
    \\
    \rotatebox{90}{\hspace{5mm}Editing output\vspace{-3mm}} &
    \includegraphics[width=30mm]{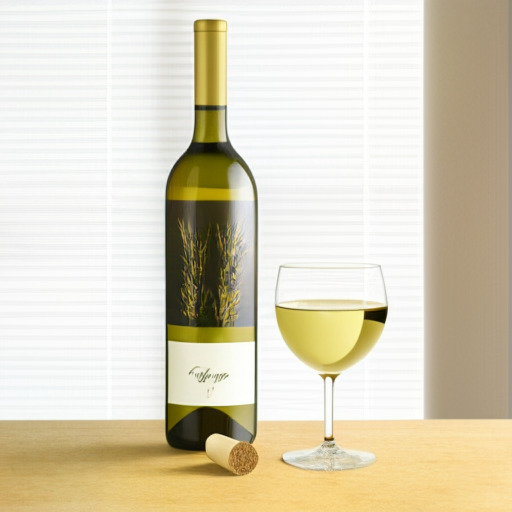} &
    \includegraphics[width=30mm]{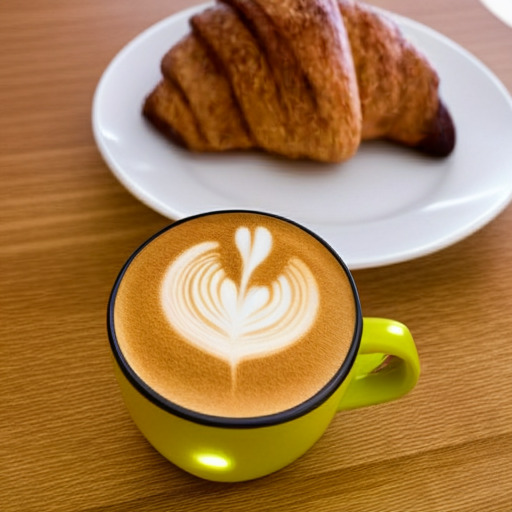} &
    \includegraphics[width=30mm]{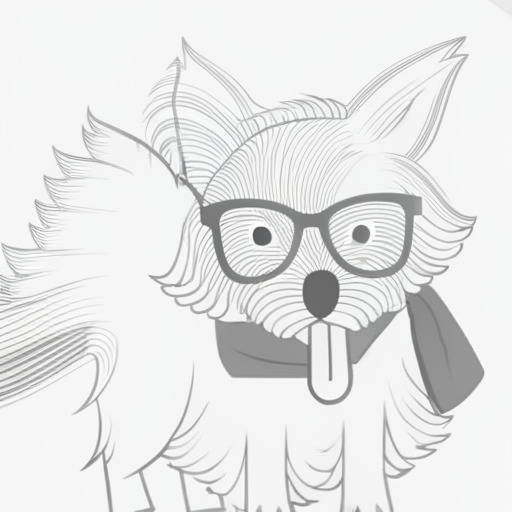} &
    \includegraphics[width=30mm]{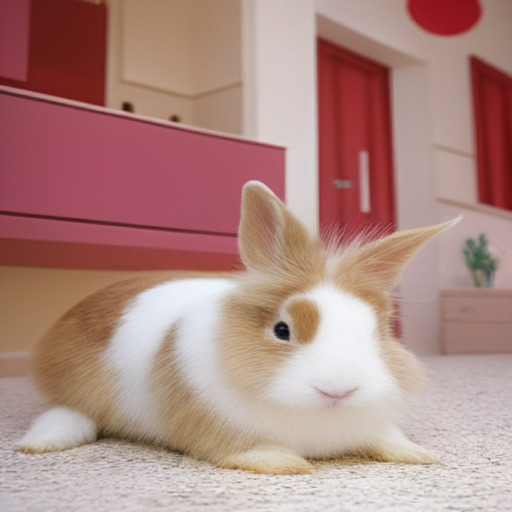} &
    \includegraphics[width=30mm]{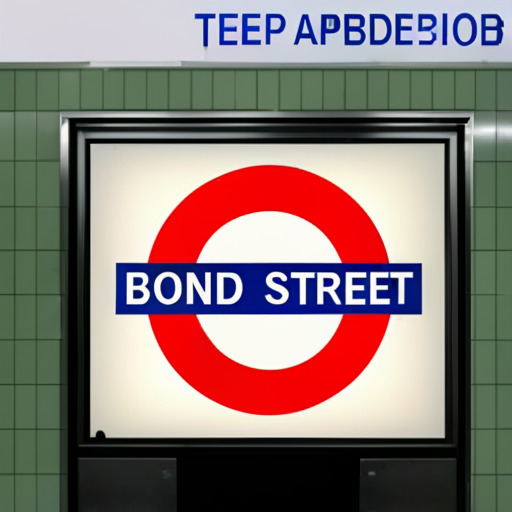} 
    \\
    &
    A bottle of Pinot Grigio next to a glass of white wine and a cork. &
    A croissant next to a latte with a flower latte art. &
    A dog. &
    A brown rabbit. &
    Bond Street.
    \\
    \rotatebox{90}{\hspace{5mm}Input image} &
    \includegraphics[width=30mm]{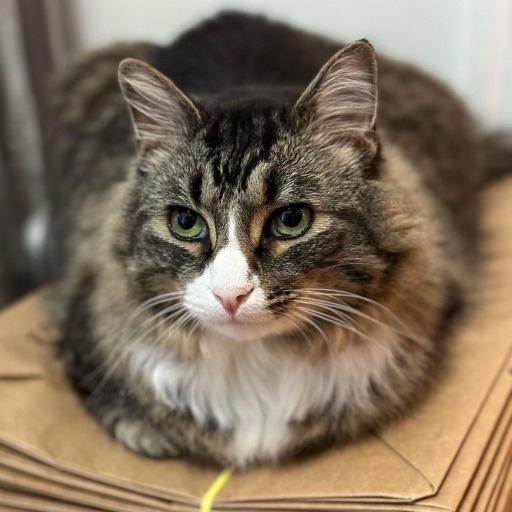} &
    \includegraphics[width=30mm]{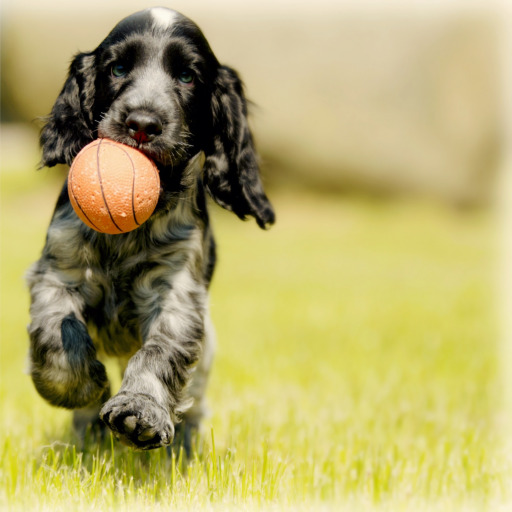} &
    \includegraphics[width=30mm]{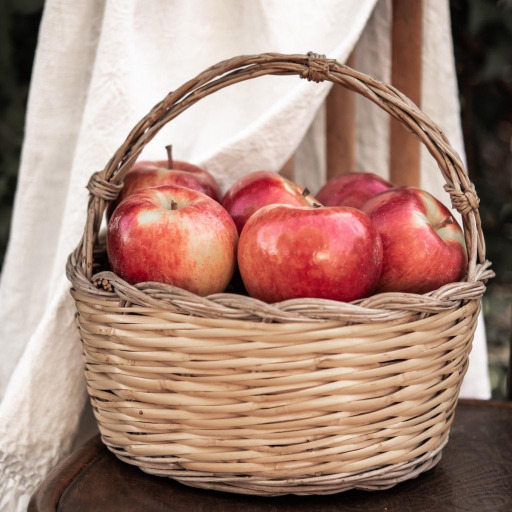} &
    \includegraphics[width=30mm]{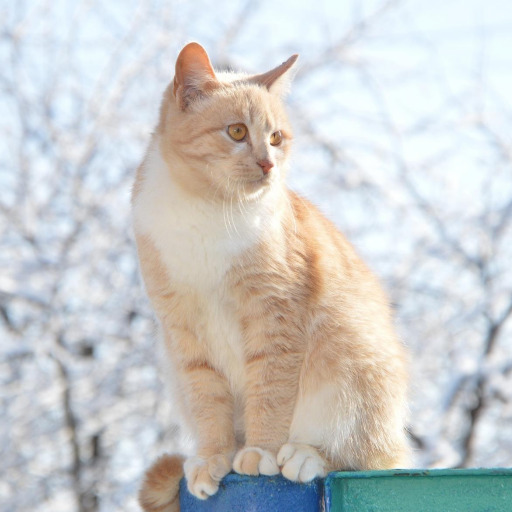} &
    \includegraphics[width=30mm]{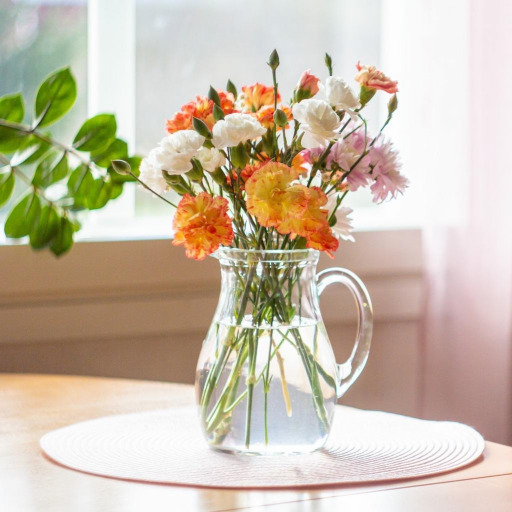}
    \\
    \rotatebox{90}{\hspace{5mm}Editing output\vspace{-3mm}} &
    \includegraphics[width=30mm]{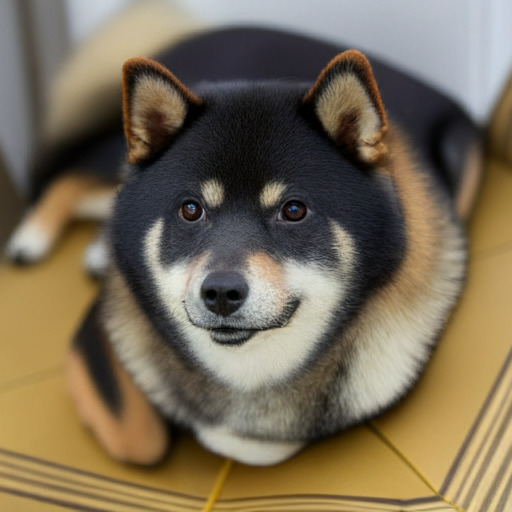} &
    \includegraphics[width=30mm]{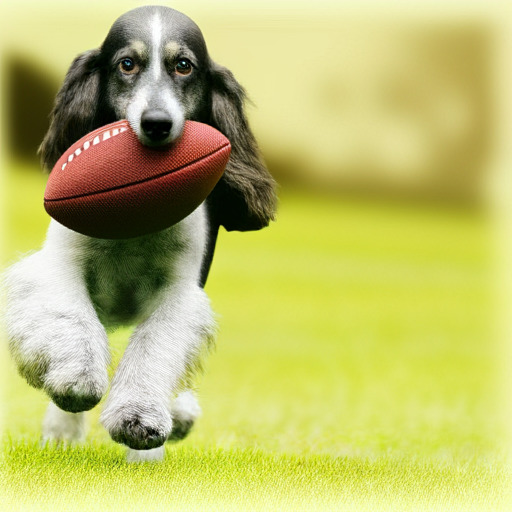} &
    \includegraphics[width=30mm]{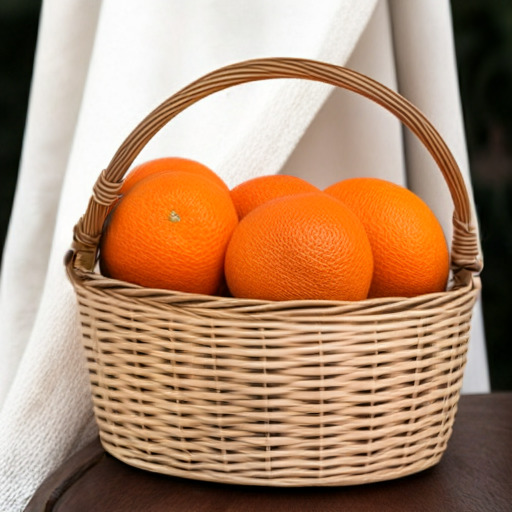} &
    \includegraphics[width=30mm]{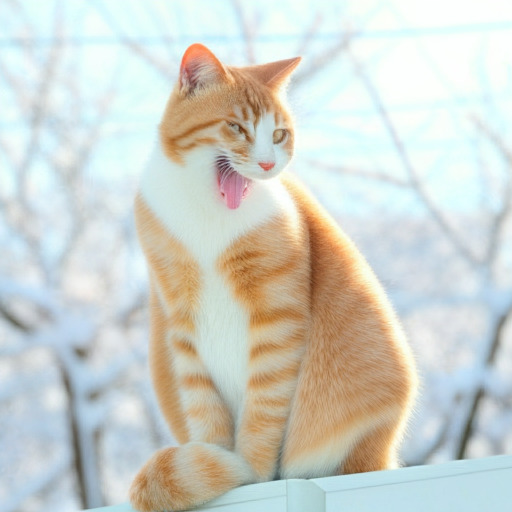} &
    \includegraphics[width=30mm]{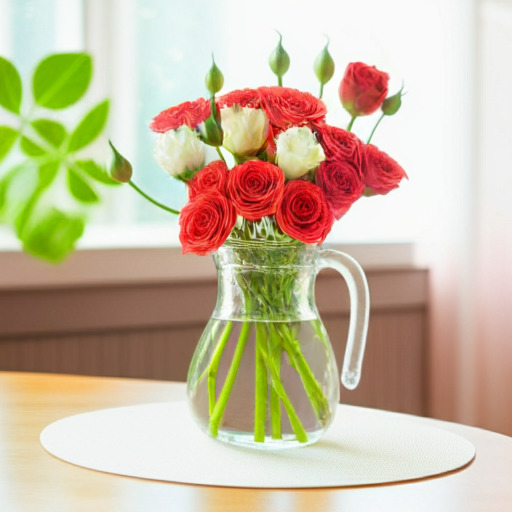} 
    \\
    &
    A Shiba Inu&
    A dog holding a football in its mouth &
    A basket of oranges&
    A photo of a cat yawning&
    A photo of a vase of red roses
  \end{tabularx}
  \caption{Examples of zero-shot mask-free image editing, post superres. We see that the pose and overall structure of the image is maintained while changing some specific aspects of the object based on the text prompt.}
  \label{fig:mfe_gallery}
\end{figure*}

\begin{figure*}
    \centering
  \begin{tabular}{p{0.95\textwidth}}
  \hspace{5mm}
    \includegraphics[width=150mm]{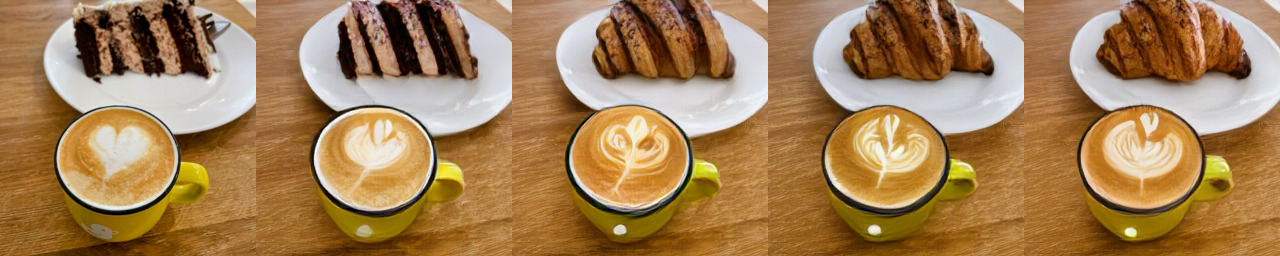}\\ 
    \hspace{5mm}
    \begin{tabularx}{150mm}{p{25mm}p{25mm}p{25mm}p{25mm}p{25mm}}
    Input  &
    Iteration 25 &
    Iteration 50 &
    Iteration 75 &
    Iteration 100
    \end{tabularx}
  \end{tabular}
    \caption{Intermediate iterations producing one of the edits in \cref{fig:mfe_gallery} (pre-superres)}
    \label{fig:edit_iter}
\end{figure*}

We perform the editing using the low-resolution base model, then perform superres on the final output (conditioned on the editing prompt). In the examples (\cref{fig:teaser_edit}, \cref{fig:mfe_gallery}), we resample 8\% of the tokens per iteration for 100 iterations, with a guidance scale of 4. We also perform top-$k$ ($k=3$) sampling on the token logits to prevent the process from diverging too much from the input. The iterative nature allows for control over the final output. \cref{fig:edit_iter} shows a few intermediate edits (without superres); in this example, the user may prefer iteration 50 or 75 over the final output.

\section{Related Work}
\subsection{Image Generation Models} 
Variational autoencoders \citep{vqvae} and Generative Adversarial Models (GANs) have shown excellent image generation performance with many variants proposed for both convolutional and Transformer architectures e.g. \citep{goodfellow2020generative,esser2021taming,karras2019style,brock2018large,donahue2019large}. Until recently, GANs were considered state of the art. Diffusion models, based on progressive denoising principles, are now able to synthesize images and video at equal or higher fidelity \citep{ddpm,kingma2021variational,ho2022video}. Hybrid approaches that combine principles from multiple approaches have also shown excellent performance \citep{maskgit,lezama2022improved}, suggesting that there are more complementarities between approaches that can be exploited.

\subsection{Image Tokenizers}
Image tokenizers are proving to be useful for multiple generative models due to the ability to move the bulk of the computation from input (pixel) space to latents \citep{ldm}, or to enabling more effective loss functions such as classification instead of regression \citep{maskgit,lezama2022improved,li2022mage}. A number of tokenization approaches such as Discrete VAE's \citep{rolfe2016discrete}, VQVAE \citep{vqvae} and VQGAN \citep{esser2021taming} have been developed, with the latter being the highest-performing as it combines perceptual and adversarial losses to achieve excellent reconstruction. ViT-VQGAN \citep{yu2021vector} extends VQGAN to the Transformer architecture. We use VQGAN rather than ViT-VQGAN as we found it to perform better for our model, noting that a better performing tokenization model does not always translate to a better performing text-to-image model.

\subsection{Large Language Models}
Our work leverages T5, a pre-trained large language model (LLM) that has been trained on multiple text-to-text tasks \citep{t5xxl}. LLMs (including T5, BERT \citep{bert}, and GPT \citep{brown2020language,radford2019language}) have been shown to learn powerful embeddings which enable few-shot transfer learning. We leverage this capacity in our model. All of the modern LLMs are trained on token prediction tasks (either autoregressive or not). The insights regarding the power of token prediction is leveraged in this work, where we apply a transformer to predict \emph{visual} tokens.

\subsection{Text-Image Models}
Leveraging paired text-image data is proving to be a powerful learning paradigm for representation learning and generative models. CLIP \citep{clip} and ALIGN \citep{jia2021scaling} train models to align pairs of text and image embeddings, showing excellent transfer and few-shot capabilities. Imagen \citep{imagen} and Parti \citep{parti} use similar large scale text-image datasets \citep{laion,schuhmann2022laion} to learn how to predict images from text inputs, achieving excellent results on FID and human evaluations. A key trick is the use of classifier-free guidance \citep{ho2022classifier,dhariwal2021diffusion} that trades off diversity and quality.

\subsection{Image Editing with Generative Models}
GANs have been extensively studied for image editing and manipulation capabilities (see \citep{xia2022gan} for a survey). A number of techniques have been developed on diffusion models to enable editing, personalization and inversion to token space \citep{gal2022image, meng2021sdedit, dreambooth, imagic, brooks2022instructpix2pix, prompttoprompt, nulltext2022}. Dreambooth \citep{dreambooth} and Imagic \citep{imagic} involve fine-tuning of the generative models. ImagenEditor \citep{imageneditor} frames the editing task as text-guided image inpainting, and involves user specified masks.

\section{Discussion and Social Impact}
The \name~model confirms the findings of \citep{imagen} that frozen large pretrained language models serve as powerful text encoders for
text-to-image generation. We also tried in our initial experiments to learn a language model from scratch on the training data, but found that performance was significantly worse than using a pre-trained LLM, especially on long prompts and rare words. We also show that non-diffusion, non-autoregressive models based on the Transformer architecture can perform at par with diffusion models while being significantly more efficient at inference time. We achieve SOTA CLIP scores, showing an excellent alignment beteween image and text. We also show the flexibility of our approach with a number of image editing applications.

We recognize that generative models have a number of applications with varied potential for impact on human society. Generative models \citep{imagen,parti,ldm,midjourney} hold significant potential to augment human creativity \citep{hughes2021generative}. However, it is well known that they can also be leveraged for misinformation, harassment and various types of social and cultural biases \citep{franks2018sex,whittaker2020all,srinivasan2021biases,steed2021image}. Due to these important considerations, we opt to not release code or a public demo at this point in time. 

Dataset biases are another important ethical consideration due to the requirement of large datasets that are mostly automatically curated. Such datasets have various potentially problematic issues such as consent and subject awareness \citep{paullada2021data, dulhanty2020issues,scheuerman2021datasets}. Many of the commonly used datasets tend to reflect negative social stereotypes and viewpoints \citep{prabhu2020large}. Thus, it is quite feasible that training on such datasets simply amplifies these biases and significant additional research is required on how to mitigate such biases, and generate datasets that are free of them: this is a very important topic \citep{buolamwini2018gender,hendricks2018women} that is out of the scope of this paper. 

Given the above considerations, we do not recommend the use of text-to-image generation models without attention to the various use cases and an understanding of the potential for harm. We especially caution against using such models for generation of people, humans and faces.

\section*{Acknowledgements}
We thank William Chan, Chitwan Saharia, and Mohammad Norouzi  for providing us training datasets, various evaluation codes and generous suggestions. Jay Yagnik, Rahul Sukthankar, Tom Duerig and David Salesin provided enthusiastic support of this project for which we are grateful.
We thank Victor Gomes and Erica Moreira for infrastructure support,  Jing Yu Koh and Jason Baldridge for dataset, model and evaluation discussions and feedback on the paper, Mike Krainin for model speedup discussions, JD Velasquez for discussions and insights, Sarah Laszlo, Kathy Meier-Hellstern, and Rachel Stigler for assisting us with the publication process, Andrew Bunner, Jordi Pont-Tuset, and Shai Noy for help on internal demos, David Fleet, Saurabh Saxena, Jiahui Yu, and Jason Baldridge for sharing Imagen and Parti speed metrics.

\bibliography{refs}
\bibliographystyle{icml2022}

\newpage
\appendix
\onecolumn
\section{Appendix.}

\subsection{Base Model Configurations}
\label{sec:base_configs}
Our base model configuration for our largest model of size 3B parameters is given in \tabb{basemodel}.

\begin{table}[ht!]
\tablestyle{6pt}{1.02}
\scriptsize
\begin{tabular}{y{107}|y{150}}
Configuration & Value \\
\shline

Number of Transformer layers & 48 \\
Transformer Hidden Dimension & 2048 \\
Transformer MLP Dimension & 8192 \\ 
Optimizer & AdaFactor \citep{shazeer2018adafactor} \\
Base learning rate & 1e-4 \\
Weight decay  & 0.045 \\
Optimizer momentum & $\beta_1{=}0.9, \beta_2{=}0.96$ \\
Batch size & 512 \\
Learning rate schedule & cosine decay \citep{Loshchilov2017SGDRSG} \\
Warmup steps & 5000 \\
Training steps & 1.5M 

\end{tabular}
\vspace{-.5em}
\caption{Configuration and training hyperparameters for base model.}
\label{tab:basemodel} \vspace{-.5em}
\end{table}

\subsection{VQGAN Configurations}
\label{sec:vqgan_configs}

\begin{table}[ht!]
\tablestyle{6pt}{1.02}
\scriptsize
\begin{tabular}{y{107}|y{150}}
Configuration & Value \\
\shline

Perceptual loss weight & 0.05 \\
Adversarial loss weight & 0.1 \\
Codebook size & 8192 \\
Optimizer & Adam \citep{KingmaB14} \\
Discriminator learning rate & 1e-4 \\
Generator learning rate & 1e-4 \\
Weight decay  & 1e-4 \\
Optimizer momentum & $\beta_1{=}0.9, \beta_2{=}0.99$ \\
Batch size & 256 \\
Learning rate schedule & cosine decay \citep{Loshchilov2017SGDRSG} \\
Warmup steps \citep{Goyal2017AccurateLM} & 10000 \\
Training steps & 1M

\end{tabular}
\vspace{-.5em}
\caption{Configuration and training hyperparameters for VQGAN.}
\label{tab:vqgan} \vspace{-.5em}
\end{table}

\textbf{VQGAN Architecture}: Our VQGAN architecture is similar to the previous work \citep{esser2021taming}. It consists of several residual blocks, downsample(encoder) and upsample (decoder) blocks. The main difference is that we remove the non-local block to make the encoder and decoder fully convolutional to support different image sizes. In the base VQGAN model, we apply 2 residual blocks in each resolution and the base channel dimension is 128. For the finetuned decoder, we apply 4 residual blocks in each resolution and we also make the base channel dimension to be 256.

\begin{figure*}[ht!]
\centering
\begin{tabularx}{0.92\textwidth}{c c c}
\includegraphics[width=0.30\textwidth]{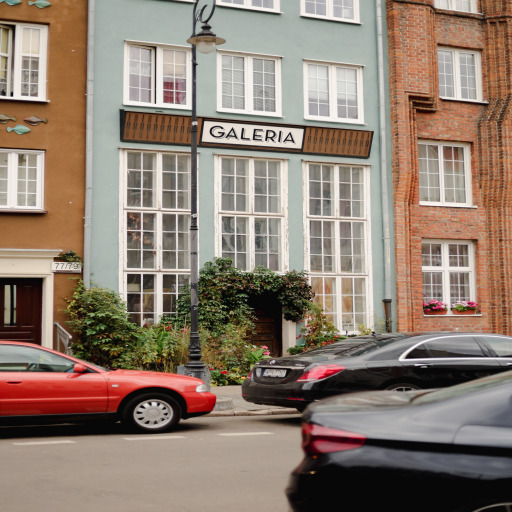} &
\includegraphics[width=0.30\textwidth]{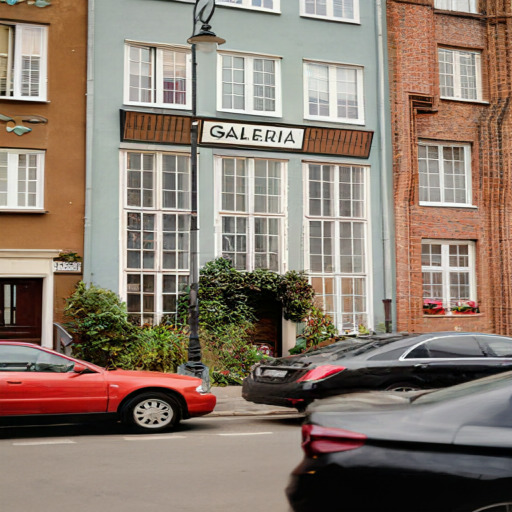} &
\includegraphics[width=0.30\textwidth]{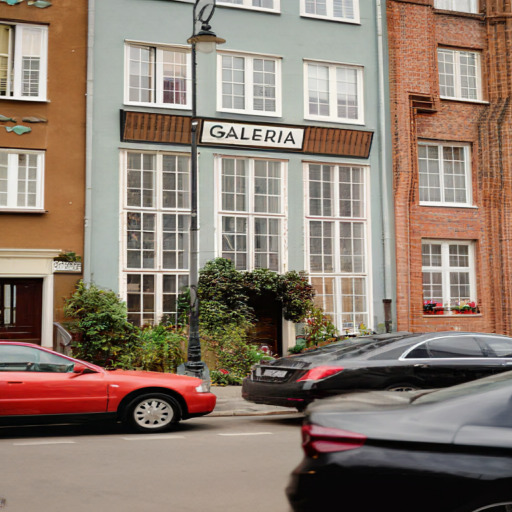}

\\
Input Image &
VQGAN Reconstruction &
Finetuned Decoder 

\end{tabularx}
\caption{\small Visual example of the improvement from the fine-tuned decoder (\cref{sec:dec_finetune}). Please zoom in by at least 200\% to see the difference between the VQGAN reconstruction and the reconstruction with a finetuned decoder. We can see especially that fine details such as the house number (bottom left), the storefront sign (middle) and the bars on the windows (right) are better preserved in the finetuned decoder.}
\label{fig:finetune_decoder}
\end{figure*}

\subsection{Super Resolution Configurations}
\label{sec:superres_configs}

\begin{table}[ht!]
\tablestyle{6pt}{1.02}
\scriptsize
\begin{tabular}{y{107}|y{150}}
Configuration & Value \\
\shline
LowRes Encoder Transformer Layers & 16 \\
Number of Transformer layers & 32 \\
Transformer Hidden Dimension & 1024 \\
Transformer MLP Dimension & 4096 \\ 
Optimizer & AdaFactor \citep{shazeer2018adafactor} \\
Base learning rate & 1e-4 \\
Weight decay  & 0.045 \\
Optimizer momentum & $\beta_1{=}0.9, \beta_2{=}0.96$ \\
Batch size & 512 \\
Learning rate schedule & cosine decay \citep{Loshchilov2017SGDRSG} \\
Warmup steps & 5000 \\
Training steps & 1M

\end{tabular}
\vspace{-.5em}
\caption{Configuration and training hyperparameters for the Super-Resolution Model.}
\label{tab:superres} \vspace{-.5em}
\end{table}

\end{document}